\documentclass{article}

\usepackage[final]{neurips_2020}

\usepackage[utf8]{inputenc} 
\usepackage[T1]{fontenc}    
\usepackage{hyperref}       
\usepackage{url}            
\usepackage{booktabs}       
\usepackage{amsfonts}       
\usepackage{nicefrac}       
\usepackage{microtype}      
\usepackage{graphicx}
\usepackage{subfig}
\usepackage{amsmath}
\usepackage{multirow}
\usepackage{soul}
\usepackage[dvipsnames]{xcolor}
\usepackage{svg}

\title{A Simple Language Model for\\ Task-Oriented Dialogue}
\author{%
Ehsan Hosseini-Asl\\
ehosseiniasl@salesforce.com \\
Salesforce Research
\And 
Bryan McCann\\
bmccann@salesforce.com \\
Salesforce Research
\AND
Chien-Sheng Wu\\
wu.jason@salesforce.com \\
Salesforce Research
\And
Semih Yavuz\\
syavuz@salesforce.com \\
Salesforce Research
\And
Richard Socher\\
rsocher@salesforce.com \\
Salesforce Research
}

\begin{document}

\maketitle

\begin{abstract}
Task-oriented dialogue is often decomposed into three tasks: 
understanding user input, 
deciding actions, and 
generating a response.
While such decomposition might suggest a dedicated model for each sub-task, we find a simple, unified approach leads to state-of-the-art performance 
on the MultiWOZ dataset.
SimpleTOD is a simple approach to task-oriented dialogue that uses a single, causal language model trained on all sub-tasks recast as a single sequence prediction problem.
This allows SimpleTOD to fully leverage transfer learning from pre-trained, open domain, causal language models such as GPT-2. 
SimpleTOD improves over the prior state-of-the-art
in joint goal accuracy for dialogue state tracking, and our analysis reveals robustness to noisy annotations in this setting.
SimpleTOD also improves the main metrics used to evaluate action decisions and response generation in an end-to-end setting:
inform rate by 8.1 points, success rate by 9.7 points, and combined score by 7.2 points. 

\end{abstract}
\section{Introduction}
\label{sec:introduction}


Conversational AI has been a long-standing area of exploration in computer science, 
and has gained more attention recently in both academia and industries 
with the current advances of neural approaches~\citep{gao2019neural}.
There are broadly two categories of dialogue. 
Open-domain dialogue systems focus on making chit-chat,
open-ended conversations with humans more natural and engaging.
They are usually trained end-to-end using large-scale data from social media~\citep{meena2020adiwardana, roller2020recipes}. 
Task-oriented dialogue (TOD) systems accomplish a goal described by a user in natural language.
They often use a pipeline approach~\citep{smith1994spoken, young2013pomdb}.
The pipeline requires natural language understanding (NLU) for belief state tracking, 
dialogue management (DM) for deciding which actions to take based on those beliefs, 
and natural language generation (NLG) for generating responses~\citep{wen2016network}.

Traditionally, each component of task-oriented dialogue systems is trained independently with different supervision. %
The NLU module is trained on domain and intent labels.
The DM module employs dialogue belief and dialogue act labels. %
The NLG module accesses templatized or natural responses.
The modular dependencies of these components can lead to error propagation when information is not provided to subsequent modules in the pipeline~\citep{liu2018end}. 
For example, many systems do not consider the entire dialogue history at every turn, 
but rather rely on the NLU module to pass belief states reliably to following module components~\citep{DAMD2019zhang}. 

We propose recasting task-oriented dialogue as a simple, causal (unidirectional) language modeling task.
We show that such an approach can solve all the sub-tasks in a unified way using multi-task maximum likelihood training. 
The proposed Simple Task-Oriented Dialogue (SimpleTOD) approach enables modeling of the inherent dependencies between the sub-tasks of task-oriented dialogue, 
by optimizing for all tasks in an end-to-end manner. 
SimpleTOD also opens the path towards fully leveraging large language models such as GPT-2~\citep{radford2019language} for task-oriented dialogue.
The success of SimpleTOD demonstrates a strong connection between the implicit language understanding in the open domain required of high-quality causal language models and the kind of understanding required for a full task-oriented dialogue system.

Evaluation results demonstrate the advantages of SimpleTOD.
It achieves 55.76 joint goal accuracy on MultiWOZ, 
which surpasses all prior work for the dialogue state tracking (i.e. belief state tracking) sub-task.
In the setting closest to testing a full task-oriented dialogue system, 
in which belief states and action decisions are generated rather than retrieved from an oracle, 
SimpleTOD performance surpasses prior work on each individual action and response generation metric (+8.1 inform rate, +9.7 success rate). 

The contributions of this work are summarized as follows:

\begin{itemize}


\item SimpleTOD --  a state-of-the-art generative model for dialogue state tracking (DST). 
\item SimpleTOD is also the first model to achieve state-of-the-art performance for dialogue state tracking, action decisions, and response generation metrics together in an end-to-end setting. 
\item Analysis showing SimpleTOD is a robust dialogue state tracker in the presence of noisy-labeled annotations. 
\item Ablations showing the importance of \textit{user/system} and \textit{endof(segment)} tokens.
\item Ablations showing the importance of pre-training that also show larger versions of SimpleTOD are not always better for end-to-end MultiWOZ.
\item A list of discovered noisy annotations in MultiWOZ 2.1 alongside a cleaned version of the test set, code for training and evaluation, are 
provided at \texttt{\url{https://github.com/salesforce/simpletod}}
\end{itemize}

\begin{figure*}[t!]
\centering
  \includegraphics[width=\linewidth]{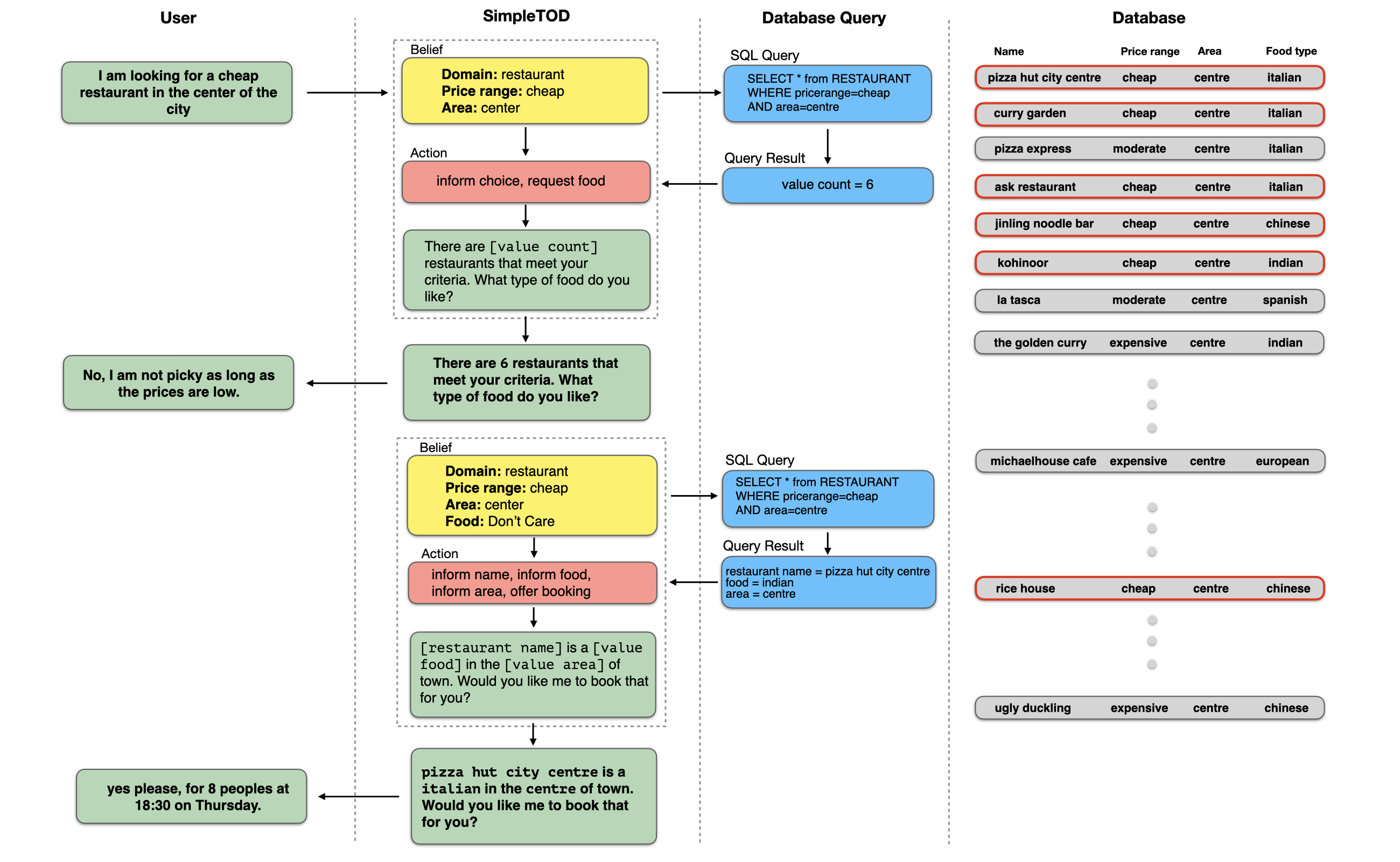}

  \caption{SimpleTOD is a simple approach to task-oriented dialogue that uses a single causal language model to generate all outputs given the dialogue context and retrieved database search results. 
  The delexicalized response can then be lexicalized into a human-readable response by using information from the belief state and DB search results.}
  \label{fig:model}
\end{figure*}

\section{Related Work}
\label{sec:related-works}

\paragraph{Task-Oriented Dialogue}
Most works on task-oriented dialogue focus on a specific module and evaluates only for that module. 
These components include understanding user intent via intent detection ~\citep{liu2016attentionbasedrnn}, 
tracking the constraints imposed by the user via dialogue state tracking~\citep{henderson2013deep, mrksic2017neuralbelief, rastogi2017scalable, nouri2018toward, trade2019wu, zhang2019find, zhou2019multi, schema2020chen, heck2020trippy}, 
determining system actions via dialogue policy~\citep{wen2017action}, 
and using dedicated response generation components~\citep{wen2015sclstm}.

Some recent works have started to bridge multiple sub-tasks by connecting modules together 
and evaluating in settings that hand off generated results from one module to another. 
~\citet{HDSA2019chen} proposed a joint action-response generation using oracle dialogue states.
~\citet{peng2020fewshot} used GPT-2 to learn a response generator conditioned on oracle dialogue acts 
which did not evaluate on dialogue state tracking.

\paragraph{Towards End-to-End Task-Oriented Dialogue}
Dependencies between these independent modules make pipeline approaches vulnerable to error propagation across components~\citep{liu2018dialogue}. 
Recent approaches have increasingly shifted towards end-to-end solutions,
which aim to reduce human effort and task-specific design. 
Several works used both dialogue history and knowledge bases as input and optimized neural encoder-decoder models to generate or retrieve system responses without modular supervision~\citep{eric2017key, zhao2017generative, madotto2018mem2seq, wu2019global, wu2019alternating}.
Some systems are mostly end-to-end, but still need to call out additional APIs or skip intermediate tasks like dialogue state tracking~\citep{bordes2017end2end}.
Others have incorporated additional supervision and trained in multi-task settings.
~\citet{lei2018sequicity} and~\citet{shu2018bsintotod} incorporated dialogue state tracking 
and jointly trained with response generation using a sequence-to-sequence approach.
~\citet{liu2018dialogue} proposed a hybrid imitation and reinforcement learning method, 
by jointly learning a policy for dialogue management with response generation.
~\citet{wen2016network, liang2019moss} trained language understanding, 
dialogue state tracking, 
and dialogue policy modules with a shared encoder.
Many other works fall somewhere in between by jointly training some tasks.
\citet{neelakantan19assistant} modeled dialogue management and response generation jointly, 
incorporating latent knowledge reasoning through attention without using belief states. 
~\citet{zhao2019rethinking} proposed to model system actions as latent variables, inducing a latent action space with variational inference methods. 
~\citet{DAMD2019zhang} proposed a domain-aware multi-decoder model (DAMD) using augmented dialogue data, 
which achieved state-of-the-art combined score for dialogue management and response generation 
on the MultiWOZ dataset.
Although all these approaches have come closer to unifying the stack, 
none are as simple as SimpleTOD: 
treating all of task-oriented dialogue as a single sequence prediction problem, 
using a single model, 
trained with a single, joint, multi-task loss.

\paragraph{Unsupervised pre-training for natural language processing}
Pre-training approaches for natural language processing focus on transferable representations 
for contextualized word vectors~\citep{mccann2017learned,peters2018deep}, 
generative models~\citep{radford2019language,keskar2019ctrl}, 
or a combination of both~\citep{dong2019unified,yang2019xlnet}.
Variants of pre-trained, bidirectional Transformers like BERT~\citep{devlin2018bert} 
are often evaluated on classification tasks such as those in the GLUE benchmark~\citep{wang2018glue} 
or span-based question answering tasks~\citep{rajpurkar2016squad}. 
Unidirectional (causal) pre-trained language models 
such as GPT-2~\citep{radford2019language} or CTRL~\citep{keskar2019ctrl} 
resemble the decoder from the original Transformer architecture~\citep{vaswani2017attention}.
They aim to learn a distribution for next-word prediction, 
which makes them particularly useful for tasks that require text generation.
In dialogue,
\citet{zhang2019dialogpt} built on GPT-2 by further pre-training it on Reddit data 
for open-domain response generation. 
\citet{henderson2019convert} also pre-trained on Reddit data with a dual Transformer encoder 
for response selection. 
\citet{bao2019plato} used both Twitter and Reddit data to pre-train a Transformer model 
with discrete latent variables. 
\citet{wu2020tod} proposed a response selection model 
by pre-training BERT model on multiple task-oriented corpora.

\citet{budzianowski2019hello} employed GPT-2 to leverage a pre-trained language model for dialogue response generation. They prepended dialogue context with belief state and DB search results. Their evaluation is only for context-to-response generation, and it is not applicable to the end-to-end setting. 
\citet{wu2019alternating} used GPT2 in an alternating roles for the user and system, without using belief state or action annotation. It achieved better results than its predecessor on context-to-response generation (dialogue policy), but it requires oracle belief states to evaluate inform and success rate which is not applicable to end-to-end evaluation. It is also not designed for dialogue state tracking.

\citet{ham2020end} fine-tuned GPT-2 on the MultiWOZ dataset and achieved lower performance on dialogue state tracking and end-to-end evaluation compared to the previous single-task and modularized models. They employed delimiter tokens for different segments: <usr>, <sys>, <ds> and <sa>. However, this differs from our tokenization, which is based on choosing semantic words which help the model to learn the semantic role of each segments -- belief state, action, and response -- and end of segment tokens such as <|endofbelief|>. They also made use of token-type embeddings for user and system sequences, whereas our model does not use this type of embedding layer. Their proposed model achieved lower performance than previous baselines for dialogue state tracking and in the end-to-end evaluation, whereas we outperform all previous models. 

\citet{peng2020soloist} proposed a GPT2-based model for end-to-end training (SOLOIST) by removing actions from the input sequence, and used token-type embedding for user and system responses as well. It is additionally pretrained on seven more dialogue datasets before fine-tuning on MultiWOZ. They used data augmentation during training using contrastive learning, similar to DAMD~\cite{DAMD2019zhang}, where they combined dialogue context and belief states with a negative response and used the final token for classification to improve end-to-end performance. However, SOLOIST did not report end-to-end performance without pretraining on other dialogue datasets. Their "w/o pretraining" setting is only evaluated in a low resource setting, and they do not evaluate on dialogue state tracking as well.

In summary, our proposed model is much simpler than previous language model based approaches, in terms of (1) input sequence definition, (2) embedding layers, (3) training algorithm, and (4) pretraining, which make it easy to reproduce the results, and outperformed previous models on DST and end-to-end setting.

\section{Methods}
\label{sec:methods}


This section describes task-oriented dialogue, how we frame it for SimpleTOD, the model architecture, training details, dataset details, and evaluation metrics.

\subsection{Task-Oriented Dialogue} 
\label{ssec:model-task}
Task-oriented dialogue (TOD) is evaluated on three sub-tasks: 
dialogue state (belief state) tracking, 
dialogue management (action/decision prediction)
and response generation. 
This decomposition has made it possible to create dedicated models for each sub-task, 
which is the dominant approach.
By contrast, 
we explore the possibility of using a single-model, 
end-to-end approach, 
SimpleTOD.

Dialogues consist of multiple turns.
In a turn $t$, the user provides input $U_t$ and the system generates a response $S_t$.
To generate a response during inference, 
SimpleTOD reads all previous turns as context, $C_t=[U_0, S_0, \ldots, U_t]$.
It generates a belief state $B_t$,
\begin{equation}
    B_t = \text{SimpleTOD}(C_t)
\end{equation}
which is a list of triplets recording values for slots in a particular domain: \textit{(domain, slot\_name, value)}. 
This belief state is used to query a database for information. The database search returns rows from the database that satisfy the conditions of the belief state. 
The rows returned can later be used to lexicalize the response (filling in generated placeholders),
but SimpleTOD only takes as input the aggregated database search results, $D_t$. 
$D_t$ includes how many rows were returned and, depending on the experimental setting, whether booking status information.
SimpleTOD then conditions on $C_t$, $B_t$, and $D_t$ concatenated together as a single sequence to decide actions, $A_t$. 
\begin{equation}
    A_t = \text{SimpleTOD}([C_t, B_t, D_t])
\end{equation}
These actions are generated as another list of triplets: \textit{(domain, action\_type, slot\_name)}.
A delexicalized response $S_t$ is generated conditioned on all prior information concatenated as a single sequence. 
\begin{equation}
    S_t = \text{SimpleTOD}([C_t, B_t, D_t, A_t])
\end{equation}
%

\begin{figure*}[htb!]
\centering
\includegraphics[width=0.76\linewidth]{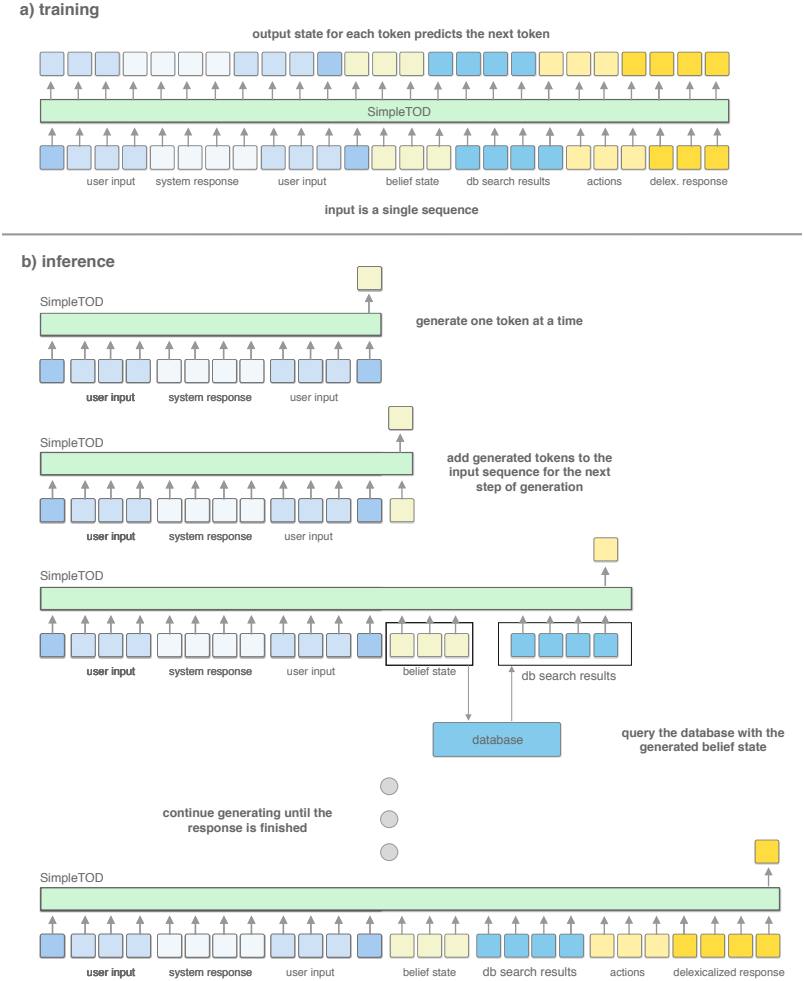}\\

  \caption{SimpleTOD is a simple approach to task-oriented dialogue that approaches all of task-oriented dialogue as a single sequence generation problem, querying a database for necessary information. 
  }
  \label{fig:simpletod}
\end{figure*}

When combined with information from the belief state and database search results, the response can be lexicalized to recover human readable response text.  Figure~\ref{fig:simpletod} depicts training of SimleTOD and generation during inference.

\subsection{Causal Language Modeling} 
\label{ssec:lm}
A single training sequence consists of the concatenation $x^t=[C_t; B_t; D_t; A_t; S_t]$
\footnote{During inference, $D_t$ comes from a database. See Sec.~\ref{sec:experimental-results} for experimental results revealing that it can be advantageous to exclude this from training.}, 
allowing us to model the joint probability over the sequence $x^t$. 
Given example sequences of the form $x=(x_1, \ldots, x_n)$ where each $x_i$ comes from a fixed set of symbols, the goal of language modeling is to learn $p(x)$. 
It is natural to factorize this distribution using the chain rule of probability~\citep{bengio2003neural}
and train a neural network with parameters $\theta$ to minimize the negative log-likelihood over a dataset $D=\{x^{1}, \ldots, x^{|D|}\}$ where sequence $x^t$ has length $n_t$: 
\begin{align}
p(x) = \prod_{i=1}^n p(x_i | x_{<i})\qquad
\mathcal{L}(D) = - \sum_{t=1}^{|D|} \sum_{i=1}^{n_t} \log {p_{\theta}(x_{i}^{t} | x^{t}_{<i})} 
\end{align}

\subsection{Architecture} 
\label{ssec:architecture}

We train a variant of the Transformer~\citep{vaswani2017attention} to learn these conditional distributions.
A sequence containing $n$ tokens is embedded as a sequence of $n$ vectors in $\mathbb{R}^d$.
Each vector is the sum of a learned token embedding and a sinusoidal positional embedding.
The sequence of vectors is stacked into a matrix $X_0\in\mathbb{R}^{n \times d}$ and processed by $l$ attention layers.
The $i$th layer consists of two blocks, 
each preserving model dimension $d$.
The first block uses multi-head attention with $k$ heads.
A causal mask precludes attending to future tokens:
\begin{align}
\text{Attention}(X, Y, Z)&=\text{softmax}\left(\frac{\text{mask}(XY^\top)}{\sqrt{d}}\right) Z\nonumber\\
\text{MultiHead}(X, k) &= [h_1;\cdots;h_k]W_o\nonumber\\
\text{where } h_j &= \text{Attention}( XW_j^1,  XW_j^2,  XW_j^3) \nonumber
\end{align}
The second block uses a feedforward network with ReLU activation that projects inputs to an inner dimension $f$. 
This operation is parameterized by $U\in \mathbb{R}^{d \times f}$ and $V\in \mathbb{R}^{f \times d}$:
\begin{align}
FF(X) = \text{max}(0, XU)V \nonumber
\end{align}
Each block precedes core functionality with layer normalization~\citep{Ba2016LayerN, child2019sparse} and follows it with a residual connection~\citep{he2016deep}. Together, they yield $X_{i+1}$:
\begin{align}
 \text{\underline{Block 1}} &&& \text{\underline{Block 2}} \nonumber
\end{align}
\begin{align}
\bar X_i &= \text{LayerNorm}(X_i) & \bar H_i &= \text{LayerNorm}(H_i) \nonumber\\
 H_{i} &= \text{MultiHead}(\bar X_i) + \bar X_i &  X_{i+1} &= \text{FF}(\bar H_i) + \bar H_i\nonumber 
\end{align}
Scores are then computed from the output of the last layer:
\begin{align*}
    \text{Scores}(X_0) = \text{LayerNorm}(X_{l}) W_{vocab}
\end{align*}
During training, these scores are the inputs of a cross-entropy loss function.
During generation, the scores corresponding to the final token are normalized with a softmax, yielding a distribution for sampling a new token.

\subsection{Training Details}
\label{ssec:train-detail}
The input to the model is tokenized with pretrained BPE codes~\citep{sennrich-etal-2016-neural} associated with DistilGPT2~\citep{sanh2019distilbert}, a distilled version of GPT-2~\citep{radford2019language}.
According to experimental results, 
Experiments for SimpleTOD use default hyperparameters for GPT-2 and DistilGPT2 in Huggingface Transformers\citep{wolf2019transformers}.
Sequences longer than $1024$ tokens are truncated.

\subsection{Dataset Details}
We evaluate on the Multi-domain Wizard-of-Oz (MultiWOZ)~\citep{budzianowski2018towards}, 
a large-scale, multi-domain dialogue dataset of human-human conversations. 
It contains 10438 multi-turn dialogues with 13.68 average turns, 
spanning over seven domains (restaurant, train, attraction, hotel, taxi, hospital, police). 
Police and hospital domains are excluded from evaluation, 
since they do not have valid/test splits.
This leaves 30 domain-slot pairs for the remaining five domain with 4,500 possible values.
SimpleTOD is trained on delexicalized system responses according to the pre-processing explained in~\cite{budzianowski2018towards}. 
Recently,~\cite{eric2019multiwoz} released MultiWOZ 2.1 which removes some noisy state values from dialogue state (belief state) tracking annotations.
For dialogue state tracking evaluation, 
we used 2.1 version in order to compare to recent state-of-the-art methods. 
To the best of our knowledge, 
all prior work on action and response generation has evaluated on 2.0, so we include those results for direct comparison.
But, we also include results for 2.1 so future work can compare to SimpleTOD on the improved version as well.

\subsection{Evaluation Details}
We follow the original MultiWOZ~\citep{budzianowski2018towards} guidance for all individual metrics and follow~\citet{mehri2019structured} for the combined score.
Joint goal accuracy is used to evaluate the performance of dialogue state tracking (i.e. belief state tracking).
It measures the accuracy of the generated belief states as they compare to oracle belief states.
Model outputs are only counted as correct when all the predicted values exactly match the oracle values.
Action and response generation uses three metrics.
The first two are inform and success rates.
They are designed to capture how well the task was completed. 
Inform rate measures how often the entities provided by the system are correct. 
Success rate refers to how often the system is able to answer all the requested attributes by user.
BLUE score ~\citep{papineni2002bleu} is used to measure the fluency of the generated responses. 
The combined score for action and response generation is computed as ($BLEU + 0.5 * (Inform + Success)$).

\section{Experimental Results and Discussion}
\label{sec:experimental-results}

\paragraph{SimpleTOD is a Unified System for Task-Oriented Dialogue} 
SimpleTOD is, to the best of our knowledge, the first system that generates state-of-the-art results judged according to dialogue state tracking as well as end-to-end metrics for action and response generation for MultiWOZ.

\subsection{Dialogue State Tracking}
\label{ssec:bst}
Table~\ref{tab:dst} compares the joint goal accuracy to previous methods.
We compare to TRADE~\citep{trade2019wu}, DSTQA~\citep{zhou2019multi}, DST-Picklist~\citep{zhang2019find}, SST~\citep{schema2020chen}, and TripPy~\cite{heck2020trippy}.
All previous models propose a bidirectional encoder to learn a better representation of the dialogue context, 
but SimpleTOD uses a unidirectional (causal) decoder and no additional bidirectional encoder.
It also makes no use of extra supervision.
It nonetheless achieves state-of-the-art. 

Many models use some form of test-label cleaning.
TRADE, DSTQA, DST-Picklist, and SST use the script proposed by~\citet{trade2019wu}\footnote{\url{https://github.com/jasonwu0731/trade-dst/blob/master/utils/fix\_label.py}}.
DSTQA also accounts for label variations that would have originally been considered incorrect.
TripPy apply their own format normalization, typo corrections, and process for accounting for label variations.
SimpleTOD achieves the best performance without any cleaning or normalization, simply on the raw, original annotations. 
Applying the script from ~\citet{trade2019wu} improves the result to 55.76.
Analysis of further noisy annotation is presented in section~\ref{sec:analysis_results}.
Further cleaning those annotations more accurately reflects performance at 57.47.
We will release the list of noisy annotations that need to be fixed along with their corrections, but we reiterate that SimpleTOD does not need this cleaning to surpass prior methods.

\begin{table}[t]
\small
\centering
\begin{tabular}{llllr}
\hline Model & Decoder & Context Encoder & Extra Supervision & Joint Accuracy \\ 
\hline
TRADE$^*$ & Generative + Classifier & Bidirectional & - & 45.6 \\ 
DSTQA$^{**}$ & Classifier & Bidirectional & knowledge graph & 51.17 \\ 
DST-Picklist$^*$ & Classifier & Bidirectional & - & 53.3 \\ 

SST$^*$ & Generative & Bidirectional & schema graph & 55.23 \\
TripPy$^{\dagger}$ & Classifier & Bidirectional & action decision & 55.3 \\
SimpleTOD$^o$ &  Generative & Unidirectional & - & 55.72 \\
SimpleTOD$^*$  &  Generative & Unidirectional & - & \textbf{55.76} \\
SimpleTOD$^+$  &  Generative & Unidirectional & - & 57.47 \\

\hline
\end{tabular}
\vspace{1mm}
\caption{\label{font-table} 
Evaluation of Dialogue State Tracking (DST) on MultiWOZ 2.1 using joint accuracy metric. 
$^*$ uses test label cleaning proposed by \citet{trade2019wu} and recommended by MultiWOZ authors. 
$^\dagger$ uses label normalization and equivalent matching proposed in \citet{heck2020trippy}. 
$^{**}$ uses the cleaning of $^*$ models plus additional accounting for label variants. 
$^+$ performs cleaning of Type 2 and partial cleaning of Type 4 noisy annotations as outlined in Section~\ref{sec:analysis_results}, which is currently non-standard and so left unbolded.
$^o$ no label-cleaning. 
}
\label{tab:dst}
\end{table}

\subsection{Action and Response Generation}
\label{ssec:action_and_response}
Table~\ref{tab:joint-task-2.0} and Table~\ref{tab:joint-task-2.1} demonstrate the effectiveness of SimpleTOD for action and response generation in the most realistic, fully end-to-end\footnote{The term "end-to-end" is overloaded in the literature. 
Evaluation that does not use oracle belief states, actions, or response is considered end-to-end even when the system itself is not trained end-to-end. 
SimpleTOD is trained end-to-end and achieves state-of-the-art in end-to-end evaluation.} setting -- when models must generate belief states, actions, and responses. 
SimpleTOD targets replacing modularized and pipelined methods that evaluate different components evaluated with oracle information.
For reference, oracle settings compare across a variety of settings against HDSA~\cite{HDSA2019chen}, ARDM~\cite{wu2019alternating}, LaRL~\cite{zhao2019rethinking}, PARG~\cite{gao2020paraphrase} can be found in the Supplementary Materials, but these comparisons are not essential for end-to-end contributions.
%
In fact, SimpleTOD is state-of-the-art in the end-to-end setting compared to the only prior work, DAMD~\citep{DAMD2019zhang}, without achieving state-of-the-art in settings that partially utilize oracle information.
This highlights that partial, oracle evaluation does not reliably transfer to the end-to-end evaluation of full systems -- only end-to-end evaluation accurately describes the performance of a full system.

Prior work uses oracle DB Search results as supervision during training and as input during inference.
We include directly comparable experiments using oracle DB Search results.
We also include experiments that completely ignore the DB Search results to show the surprising effectiveness of SimpleTOD without DB Search information.
We also show a setting with dynamic DB Search results.
In this setting, 
we train with the number of matched DB entries and compute this dynamically at inference from generated belief states.
In all variations,
SimpleTOD outperforms prior work.

DAMD~\cite{DAMD2019zhang} is the only prior work that has evaluated with generated belief states from dialogue state tracking during inference.
We found in additional ablation experiments that we could increase scores for individual metrics like inform rate and success rate by training three separate SimpleTOD language models: 
one for dialogue state tracking, one for action generation, and one for response generation.
However, the combined scores remained nearly identical to the full end-to-end, single model approach.
For example, separating the models might improve inform rate, but hurt response generation measured by BLEU.
Regardless, in this most realistic setting
SimpleTOD achieves state-of-the-art on inform and success metric. 
SimpleTOD performs lower only on BLEU by 1.59 points, perhaps due to lack of action/response augmentation employed by DAMD.  

\paragraph{Regarding Oracle DB Search Results}
In the case where we dynamically compute partial DB Search results (number of entries matched only), the results are actually lower than ignoring them entirely.
Using oracle DB information likewise leads to lower performance.
The best result ignores DB Search results entirely.
We have found that in some cases, the generated belief states conflict in some way with the information in the database.
For example, there can be discrepancies between the two in the name of restaurants: `pizza hut fenditton' in the target belief states but `pizza hut fen ditton' in the database.
We have consulted with the authors of the dataset,
but there is currently no course of action planned to remedy this.

\begin{table*}[t!]
\centering
\small
\begin{tabular}{llllrrrr}
\hline
Model & Belief State & DB Search & Action & Inform &Success & BLEU & Combined\\
\hline
DAMD+augmentation & generated & oracle  & generated & 76.3 & 60.4 & 16.6 & 85 \\
SimpleTOD (ours) & generated & oracle & generated & 78.1 & 63.4 & 16.91 & 87.66 \\
SimpleTOD (ours) & generated & dynamic & generated & 81.4 & 69.7 & 16.11 & 91.66 \\
SimpleTOD (ours) & generated & - & generated & \textbf{84.4} & \textbf{70.1} & 15.01 & \textbf{92.26} \\
\hline
\end{tabular}
\caption{\label{tab:joint-task-2.0}
Action and response generation on MultiWOZ 2.0 reveals that SimpleTOD, a single, causal language model, is sufficient to surpass prior work. 
}
\end{table*}

\begin{table*}[t!]
\centering
\small
\begin{tabular}{lllrrrr}
\hline
  Belief State & DB Search & Action & Inform &Success & BLEU & Combined\\
\hline
 generated & oracle & generated & 79.3 & 65.4 & 16.01 & 87.36 \\
generated & dynamic & generated & 83.4 & 67.1 & 14.99 & 90.24 \\
 generated & - & generated & 85 & 70.5 & 15.23 & 92.98 \\
\hline
\end{tabular}
\caption{\label{tab:joint-task-2.1}
Action and response generation on MultiWOZ 2.1 for SimpleTOD. 
}
\end{table*}
 
\section{Analysis and Further Discussion}
\label{sec:analysis_results}

\paragraph{The Role of Special Tokens} 
Table~\ref{tab:tokenization-ablation}  evaluates SimpleTOD with different special tokens used to identify components of the input corresponding to different sub-tasks.
Analysis revealed that without end tokens, 
SimpleTOD tended to generate much longer belief state, action, and response generations. 
Even more important is clearly differentiating user and system text for SimpleTOD.

\paragraph{Pre-training}
Table~\ref{tab:pretraining-ablation} highlights the importance of initializing SimpleTOD with pre-trained weights. 
A major advantage of recasting as single sequence prediction is the ability to leverage the understanding learned by these pre-trained models in the open-domain setting.

\begin{table*}[t]
\centering
\small
\begin{tabular}{llrrrrr}
\hline
 End token & User/System token & Joint Acc & Inform& Success& BLEU & Combined\\

\hline
 No & No  & 16.79 & 33.8 & 10.6 & 4.53 & 26.73 \\
   Yes & No  & 21.5 & 54.5 & 41.2 &  9.48 & 57.33 \\
   No & Yes  & 22.22 & 61.9 & 52.7 & 9.57 & 66.87 \\
   Yes & Yes  & 55.76 & 85 & 70.5 & 15.23 & 92.98 \\
\hline
\end{tabular}
\caption{\label{tab:tokenization-ablation}
Ablations on MultiWOZ 2.1 comparing the presence and absence of different special tokens when representing TOD as a single sequence. 
Performance on all metrics drops without \textit{<endof(segment)>} and \textit{<user/system>} tokens.
}
\end{table*}

\begin{table*}[t]
\centering
\small
\begin{tabular}{llrrrrr}
\hline
Layers & Pretrained & Joint Acc & Inform& Success& BLEU & Combined\\
\hline
 6& Random  & 16.45 & 63.5 & 49.6 & 6.34 & 62.89 \\
6& DistilGPT2  & 54.54 & 85 & 70.5 & 15.23  & 92.98 \\
12& Random & 20.17 & 58.7 & 37.4 & 8.9 & 59.65 \\
12 & GPT2 & 55.76  & 88 & 61.7 & 15.9 & 90.75 \\
\hline
\end{tabular}
\caption{Ablations  on MultiWOZ 2.1 comparing the importance of pretraining. Recasting as single sequence prediction enables fully leveraging pre-trained models for the language understanding they have gathered in an open-domain setting. }
\label{tab:pretraining-ablation}
\end{table*}

\paragraph{Robustness to Noisy Annotations}
To understand the source of dialogue state tracking errors, 
we investigated MultiWOZ 2.1 annotations in depth.
In the process, 
we have defined four primary types of noisy-labels that could be considered mis-annotations:
\begin{enumerate}
    \item User provided multiple options, but context does not provide sufficient information to determine the true belief state.
    \item Belief state is not labeled, but context provides sufficient information.
    \item Belief state is labeled, but context lacks necessary information.
    \item Belief state value is misspelled according to the context information.
\end{enumerate}

Together experimental results and this analysis indicate that SimpleTOD can track dialogue state and generate the correct output even in the presence of noisy labels.
Concrete examples of noisy-labeled annotation in MultiWOZ can be found in the Supplementary Materials.
All mis-annotated examples along with all code for replication are  provided~\footnote{\url{https://github.com/salesforce/simpletod}}. 





\paragraph{Decoding}
Initialized from pre-trained weights, 
SimpleTOD does not need to employ an advanced, more costly decoding strategy such as beam search, diverse beam search, and top-k sampling as opposed to HDSA~\cite{HDSA2019chen} and DAMD~\cite{DAMD2019zhang}.
Our results are reported with simple greedy decoding.
In initial experiments,
we also tried nucleus sampling~\citep{holtzman2019curious}, 
but we found it degraded performance.
This relates to the observations in~\citet{keskar2019ctrl} around controllable generation: when precision is required, sampling from the distribution is inherently less reliable than greedily sampling.

\paragraph{Full Dialogues, Multiple Turns, and Long Contexts} In further analysis, we found that SimpleTOD accurately tracks dialogue state over multiple turns and long contexts. 
In some cases, earlier belief state errors are remedied later on when additional turns provide increased context.
Examples of full dialogues and those with many turns or especially long context can be found in Supplementary Materials, but we do not consider this further analysis as a primary contribution listed for the work.

\section{Conclusion}
\label{sec:conclusion}
We explored a simple approach to task-oriented dialogue (SimpleTOD) that uses a single, causal language model. 
To do this, during training we treat all inputs for dialogue state tracking, action and response generation as a single sequence to the model.
SimpleTOD can then directly leverage pre-trained models like GPT-2 to transfer language understanding from open-domain settings where data is more readily available.
Empirical results on the multi-domain dialogue dataset (MultiWOZ) showed that the proposed approach outperformed all prior methods in dialogue state tracking as well as in action and response generation in the end-to-end setting. 
We found that the pre-trained weights were essential, but to leverage these weights fully we had to guide the system with special tokens that mark user and system responses as well as different portions of the sequence related to different sub-tasks.
We found that SimpleTOD was effective at tracking dialogue state over long context with many turns and required no more than greedy decoding to achieve new state-of-the-art results despite noisy annotations.
We hope that these results and the code, models, and discovered noisy annotations 
will encourage further exploration of simple, unified approaches for dialogue systems.

\section{Broader Impact}
\label{sec:broader_impact}
This work may have implications for the simplification of conversational agents. 
In the narrow sense, this work addresses task-oriented dialogue,
but similar results might also hold for open-domain conversational systems.
If so, the improvement of these systems and easier deployment would amplify both the positive and negative aspects of conversational AI.
Positively, conversational agents might play a role in automating predictable communications, thereby increasing efficiency in areas of society that currently lose time navigating the multitude of APIs, webpages, and telephonic systems that are used to achieve goals.
Negatively, putting conversational agents at the forefront might dehumanize communication that can be automated and might lead to frustration where human agents could provide more efficient solutions -- for example, when predicted solutions do not apply.
These consequences are not specific to this work, but should be considered by the field of conversational AI more broadly.
\small
\bibliography{sections/6-references}
\newpage


\appendix

\section{Input Representation and Method Overview}
\label{appendix:input_rep}
As described in Section~\ref{sec:methods},
a single training sequence consists of the concatenation of context $C_{t}$, belief states $B_{t}$, database search results $D_{t}$, action decisions $A_{t}$, and system response $S_{t}$. 
A schematic overview of each segment is shown in Table~\ref{tab:method} together with special tokens marking transition points. 
SimpleTOD is optimized by minimizing the negative likelihood over the joint sequence $x^t=[C_t; B_t; D_t; A_t; S_t]$. 
The output state associated with each input token is used to predict the next token, see Figure~\ref{fig:model}a. 

During inference, SimpleTOD generates this sequence token by token, but we stop after belief states are generated to query from a database.
The outputs of the database are summarized and concatenated to the end of the input sequence and generation resumes token by token.
This results in a delexicalized response, see Figure~\ref{fig:model}b.
This response can then be lexicalized by replacing slots and values with information from the database results.
This process is described more formally in the equations of Section~\ref{tab:method}. 

\begin{table}[htb!]
\small
\centering
\begin{tabular}{|l|l|}
\hline
Context & 
    \textcolor{blue}{[context]} 
    \textcolor{TealBlue}{[user]} 
    \textit{user input}
    \textcolor{Periwinkle}{[system]} 
    \textit{system response}
    $\ldots$
    \textcolor{TealBlue}{[user]} 
    \textit{user input} 
    \textcolor{blue}{[endofcontext]}
    \\& \\
Belief State &
 \textcolor{green}{[belief]}
 \textit{domain slot\_name value, domain slot\_name value, $\ldots$}
 \textcolor{green}{[endofbelief]}
 \\
 & \\
DB Search &  
\textcolor{purple}{[db]}
 \textit{\#\_matches, booking\_status}
 \textcolor{purple}{[endofdb]}
 \\& \\
Action & 
\textcolor{orange}{[action]}
 \textit{domain action\_type slot\_name, domain action\_type slot\_name, $\ldots$}  
 \textcolor{orange}{[endofaction]}
 \\& \\
Response &  
\textcolor{red}{[response]}
 \textit{system delexicalized response}
 \textcolor{red}{[endofresponse]}
 \\
\hline
\end{tabular}
\vspace{1mm}
\caption{A schematic representation of the different components of inputs/outputs in task-oriented dialogue. When training SimpleTOD, these are concatenated together into a single sequence.}
\label{tab:method}
\end{table}





\clearpage

\section{SimpleTOD with Oracle information}
\label{appendix:multiwoz_2.1}
This section reports the performance of SimpleTOD for action and response generation, in the presence of different oracle information, i.e. oracle belied and oracle action.
These settings are not end-to-end as in the main text, and SimpleTOD is designed to be end-to-end.
We report results in these settings for a complete understanding of SimpleTOD,
but we note that only the end-to-end settings in the main text evaluate the full system together. 
In these oracle settings, 
other methods can outperform SimpleTOD, 
but this simply highlights the importance of end-to-end evaluation: there is a disconnect between performance with oracle information and performance without it. 
In practical use, oracle information is not available, and that is where SimpleTOD excels.

We report results in~Table~\ref{tab:joint-task} for two different settings regularly employed in the literature.
These settings are determined by how much oracle information is used.
The first setting uses oracle belief states and oracle actions.
The second uses oracle belief states, but requires the system to generate its own actions.

Note that all prior works use oracle DB Search results as supervision during training and as input during inference in all these settings.
We include directly comparable experiments using oracle DB Search results for all settings.
We also include experiments that completely ignore the DB Search results in all settings to show the surprising effectiveness of SimpleTOD without DB Search results.

 The evaluation results on MultiWOZ 2.1, as shown in Table~\ref{tab:appendix-joint-task-2.1}, also follow the same patterns as discussed in section~\ref{ssec:action_and_response}.
 We provide these results for future comparisons on the improved version of the dataset.

\begin{table*}[htb!]
\centering
\small
\begin{tabular}{llllrrrr}
\hline
Model & Belief State & DB Search & Action & Inform &Success & BLEU & Combined\\
\hline
DAMD+augmentation & oracle & oracle & oracle & \textbf{95.4} & \textbf{87.2} & \textbf{27.3} & 118.5 \\
PARG & oracle & oracle & oracle & 91.1 & 78.9 & 18.8 & 103.8 \\
SimpleTOD (ours) & oracle & oracle & oracle & 93.4 & 83.2 & 17.78 & 106.08 \\
SimpleTOD (ours) & oracle & - & oracle & 92.3 & 85.8 & 18.61 & 107.66 \\
\hline
HDSA & oracle & oracle & generated & 82.9 & 68.9 & 23.6 & 99.5 \\
DAMD+augmentation & oracle & oracle & generated & \textbf{89.2} & 77.9 & 18.6 & 102.5 \\
ARDM & oracle & oracle & - & 87.4 & 72.8 & 20.6 & 100.7 \\
LaRL & oracle & oracle & generated & 82.78 & \textbf{79.2} & 12.8 & 93.79 \\
SimpleTOD (ours) & oracle & oracle & generated & 84 & 72.8 & 16.1 & 94.5 \\
SimpleTOD (ours) & oracle & - & generated & 88.9 & 67.1 & 16.9 & 94.9 \\
\hline
\end{tabular}
\caption{\label{tab:joint-task}
SimpleTOD results on MultiWOZ 2.0 using oracle information.
}
\end{table*}

\begin{table*}[htb!]
\centering
\small
\begin{tabular}{lllrrrr}
\hline
  Belief State & DB Search & Action & Inform &Success & BLEU & Combined\\
\hline

 oracle & oracle & oracle & 92.8 & 84.5 & 18.9 & 107.55 \\

  oracle & - & oracle & 92.6 & 86.1 & 17.67 & 107.2 \\
  \hline

 oracle & oracle & generated & 85.1 & 73.5 & 16.22 & 95.52 \\

  oracle & - & generated & 89.6 & 68.6 & 15.46 & 94.56 \\
  


\hline
\end{tabular}
\caption{\label{tab:appendix-joint-task-2.1}
SimpleTOD results on MultiWOZ 2.1 using oracle information.
}
\end{table*}

\newpage


\section{Dialogue State Tracking Analysis}
\label{appendix:dst_analysis}

This section provides more detailed analysis of Dialogue State Tracking (DST) task performance by SimpleTOD, as mentioned in section~\ref{sec:analysis_results}. 

\paragraph{Understanding Long, Multi-domain Context} Table~\ref{tab:dst_example} indicates the DST performance of SimpleTOD in case of multi-domain and long context. This example also shows understanding dialogue states, where slots across domains are related. For example, \textit{taxi departure} and \textit{taxi destination} should be inferred from \textit{attraction name} and \textit{hotel name}.

\begin{table}[htb!]
    \centering
    \scriptsize
    \begin{tabular}{c|p{10cm}}
    \specialrule{.3em}{.2em}{.2em}
        \multirow{18}{*}{\textbf{Context ('MUL1015', turn 10)}} & { \textcolor{blue}{<|context|>} \textcolor{TealBlue}{<|user|>} i am looking for a pool somewhere in the south of cambridge . \textcolor{Periwinkle}{<|system|>} i am sorry , but it does not look like we have a pool that matches your criteria . \textcolor{TealBlue}{<|user|>} how about some type of theater ? \textcolor{Periwinkle}{<|system|>} \hl{the junction theatre} is in the south area . would you like the address ? \textcolor{TealBlue}{<|user|>} yes , please provide the address and postcode as well . \textcolor{Periwinkle}{<|system|>} the junction theatre is located on clifton way at postcode cb17gx . is there anything else i can assist you with ? \textcolor{TealBlue}{<|user|>} thanks . yes , i also need to find a guesthouse for my stay there . \textcolor{Periwinkle}{<|system|>} are you needing the guesthouse in the south , also ? do you have a preference in price range ? \textcolor{TealBlue}{<|user|>} the area should be the centre . i need it to be cheap and have free parking . \textcolor{Periwinkle}{<|system|>} i have 2 guesthouses that fit your criteria . 1 is the alexander bed and breakfast and the other is el shaddai . \textcolor{TealBlue}{<|user|>} what is the star rating -s for them ? \textcolor{Periwinkle}{<|system|>} the alexander bed and breakfast has a 4 star rating while the el shaddai has 0 stars . would you like to book a room with either of these ? \textcolor{TealBlue}{<|user|>} ok great , thank you . i also need to book a taxi to get back and forth . \textcolor{Periwinkle}{<|system|>} where did you wan to leave from and go to ? \textcolor{TealBlue}{<|user|>} i'll be leaving the theatre at 14:45 and going to the guesthouse . \textcolor{Periwinkle}{<|system|>} let s back up a moment . which lodgings did you prefer , so we know where the taxi will be taking you ? \textcolor{TealBlue}{<|user|>} \hl{let s go with the alexander bed and breakfast .} \textcolor{Periwinkle}{<|system|>} your booking is complete , a grey volkswagen will be picking you up , contact number 07753429052 . can i help with anything else ? \textcolor{TealBlue}{<|user|>} cool beans ! sorry i was so much trouble . thanks so much for all of your help , you need a raise dealing with customers like me ! \textcolor{blue}{<|endofcontext|>}}
 \\
        \hline
        \multirow{5}{*}{\textbf{Target}}  & \textcolor{green}{<|belief|>} attraction area south, attraction name the junction, attraction type theatre, hotel area centre, hotel name alexander bed and breakfast, hotel parking yes, hotel pricerange cheap, hotel type guesthouse, taxi departure the junction, taxi destination alexander bed and breakfast, taxi leaveat 14:45 \textcolor{green}{<|belief|>} \textcolor{brown}{<|action|>} general bye none \textcolor{brown}{<|endofaction|>} \textcolor{purple}{<|response|>} it was no trouble at all ! i hope you have a fantastic trip and enjoy your stay ! \textcolor{purple}{<|endofresponse|>}
 \\
        \hline
        \multirow{5}{*}{\textbf{SimpleTOD}} & \textcolor{green}{<|belief|>} attraction area south, attraction name the junction, attraction type theatre, hotel area centre, hotel name alexander bed and breakfast, hotel parking yes, hotel pricerange cheap, hotel type guesthouse, \hl{taxi departure the junction}, \hl{taxi destination alexander bed and breakfast}, taxi leaveat 14:45 \textcolor{green}{<|endofbelief|>} \textcolor{brown}{<|action|>} general bye none \textcolor{brown}{<|endofaction|>} \textcolor{purple}{<|response|>} thank you for using our service \textcolor{purple}{<|endofresponse|>}
\\
    \specialrule{.3em}{.2em}{.2em}
    \end{tabular}
    \vspace{1.5mm}
    \caption{Example of multi-domain dialogue state tracking for long context, where dialogue context spanning over 3 domains, i.e. attraction, hotel, and taxi. The belief states for taxi departure and destination (highlighted) should be inferred from hotel and attraction domains.}
    \label{tab:dst_example}
\end{table}




\paragraph{Regarding Noisy Annotations} 

Concrete examples of four types of noisy-labeled annotation defined in section~\ref{sec:analysis_results}, are shown in Tables~\ref{tab:appendix-type1-annotation},~\ref{tab:appendix-type2-annotation-1},~\ref{tab:appendix-type2-annotation-2},~\ref{tab:appendix-type3-annotation} and~\ref{tab:appendix-type4-annotation}.
The results indicate that SimpleTOD is robust to noisy annotation and can often generate the correct belief state in situations where annotation is incorrect.
As mentioned in section~\ref{ssec:bst}, 
the list of noisy annotations is released with this paper.

\paragraph{Full Dialogue Example}
Table~\ref{tab:target_response_example} and Table~\ref{tab:target_response_example_lex} include several turns as part of a full dialogue for delexicalized and lexicalized outputs.
These are from our strongest SimpleTOD model in the setting that requires it to generate all outputs, which means that it ignores all DB Search results.


\begin{table}[htb!]
    \centering
    \scriptsize
    \begin{tabular}{c|p{10cm}}
    \specialrule{.3em}{.2em}{.2em}

    \multirow{4}{*}{\textbf{Context ('SNG01608', turn 2)}} & {\textcolor{blue}{<|context|>} \textcolor{TealBlue}{<|user|>}  are there any portuguese restaurant -s in cambridge ? \textcolor{Periwinkle}{<|system|>} yes there is a portuguese restaurant in cambridge with 2 different location -s , would you like the addresses ? \textcolor{TealBlue}{<|user|>} \hl{if 1 of them has a moderate price range please give me that address . if not tell me about turkish restaurant -s instead .}  \textcolor{blue}{<|endofcontext|>}}
 \\
        \hline
        \multirow{1}{*}{\textbf{Target}}  & \textcolor{green}{<|belief|>} 
        \hl{restaurant food turkish}, restaurant pricerange moderate
        \textcolor{green}{<|endofbelief|>} 
 \\
        \hline
        \multirow{1}{*}{\textbf{SimpleTOD}} & \textcolor{green}{<|belief|>} 
        \hl{restaurant food portugese}, restaurant pricerange moderate\textcolor{green}{<|endofbelief|>} 
\\
    \specialrule{.3em}{.2em}{.2em}
    
    \multirow{5}{*}{\textbf{Context ('MUL1575', turn 3)}} & {\textcolor{blue}{<|context|>} \textcolor{TealBlue}{<|user|>} i would like to find a particular restaurant in cambridge . the name of the restaurant is restaurant 2 two . could you give me the location ? \textcolor{Periwinkle}{<|system|>} restaurant 2 two is nice french restaurant located at 22 chesterton road chesterton . would like me to book you a table ? \textcolor{TealBlue}{<|user|>} that would be great . i need it for 8 on friday . \textcolor{Periwinkle}{<|system|>} do you have a time preference ? \textcolor{TealBlue}{<|user|>} \hl{yes at 11:15 if that s not available i can do 10:15} \textcolor{blue}{<|endofcontext|>}}
 \\
        \hline
        \multirow{2}{*}{\textbf{Target}}  & \textcolor{green}{<|belief|>} restaurant book day friday, restaurant book people 8, \hl{restaurant book time 10:15}, restaurant name restaurant 2 two\textcolor{green}{<|endofbelief|>} 
 \\
        \hline
        
        \multirow{2}{*}{\textbf{SimpleTOD}} & \textcolor{green}{<|belief|>} restaurant book day friday, restaurant book people 8, \hl{restaurant book time 11:15}, restaurant name restaurant 2 two \textcolor{green}{<|endofbelief|>} 
\\
    \specialrule{.3em}{.2em}{.2em}

         

     \multirow{8}{*}{\textbf{Context ('PMUL1194', turn 6)}} & {\textcolor{blue}{<|context|>} \textcolor{TealBlue}{<|user|>}   
    hello , i need information about old schools . \textcolor{Periwinkle}{<|system|>} there are 5 local colleges , do you have a particular area you are interested in ? \textcolor{TealBlue}{<|user|>} i do not have an area preference . \textcolor{Periwinkle}{<|system|>} old schools is an architectural attraction in the center of town . they are located on trinity lane , in postcode cb21tt . \textcolor{TealBlue}{<|user|>} what is the phone number ? \textcolor{Periwinkle}{<|system|>} the phone number is 01223332320 . do you need any more information ? \textcolor{TealBlue}{<|user|>} yes , i am looking for \hl{a train that leaves on wednesday for cambridge} \textcolor{Periwinkle}{<|system|>} i have over 200 entries . let s try to narrow that down . what time would you like to arrive ? \textcolor{TealBlue}{<|user|>} \hl{monday at 11:00 for 5 people please} \textcolor{Periwinkle}{<|system|>} just to confirm , \hl{you want to leave on monday , or wednesday ?} and where are you departing from ? \textcolor{TealBlue}{<|user|>} leaving ely anytime after 15:15 and i need to know how long the ride is and the specific arrival time please .
    \textcolor{blue}{<|endofcontext|>}}
 \\
        \hline
        \multirow{2}{*}{\textbf{Target}}  & \textcolor{green}{<|belief|>} 
    attraction name old schools, train book people 5, \hl{train day monday}, train departure ely, train destination cambridge, train leaveat 15:15
        \textcolor{green}{<|endofbelief|>} 0
 \\
        \hline
        \multirow{1}{*}{\textbf{SimpleTOD}} & \textcolor{green}{<|belief|>}
    attraction name old schools, train book people 5, \hl{train day wednesday}, train departure ely, train destination cambridge, train leaveat 15:15
         \textcolor{green}{<|endofbelief|>} \\
         
     \specialrule{.3em}{.2em}{.2em}
     
       \multirow{12}{*}{\textbf{Context ('MUL1024', turn 7)}} & {\textcolor{blue}{<|context|>} \textcolor{TealBlue}{<|user|>}   
    i am looking for a theatre to visit in the centre of cambridge . can you help me with this ? \textcolor{Periwinkle}{<|system|>} i can recommend the adc theatre on park street or the cambridge arts theatre located at 6 st edward s passage \textcolor{TealBlue}{<|user|>} are either in the city center ? \textcolor{Periwinkle}{<|system|>} both are located in the city centre . \textcolor{TealBlue}{<|user|>} can you give me the address of the adc theatre please ? \textcolor{Periwinkle}{<|system|>} the address is park street . \textcolor{TealBlue}{<|user|>} what films are playing today ? \textcolor{Periwinkle}{<|system|>} i am sorry . i do not have that information . can i help you with anything else ? \textcolor{TealBlue}{<|user|>} i am looking for a particular hotel called avalon . \textcolor{Periwinkle}{<|system|>} avalon is located at 62 gilbert road . would you like me to book a reservation ? \textcolor{TealBlue}{<|user|>} great , can i book that for 6 people for 3 nights starting on tuesday . \textcolor{Periwinkle}{<|system|>} i am sorry , the booking was unsuccessful . would you like to try for another date or a shorter stay ? \textcolor{TealBlue}{<|user|>} would you be interested in \hl{staying at the avalon for 1 night ? if not we can find another hotel available for a 3 night stay for 6 guests} .
    \textcolor{blue}{<|endofcontext|>}}
 \\
        \hline
        \multirow{2}{*}{\textbf{Target}}  & \textcolor{green}{<|belief|>} 
    attraction area centre, attraction name adc theatre, attraction type theatre, hotel book day tuesday, hotel book people 6, \hl{hotel book stay 3}, hotel name avalon
        \textcolor{green}{<|endofbelief|>} 
 \\
        \hline
        \multirow{2}{*}{\textbf{SimpleTOD}} & \textcolor{green}{<|belief|>}
    attraction area centre, attraction name adc theatre, attraction type theatre, hotel book day tuesday, hotel book people 6, \hl{hotel book stay 1}, hotel name avalon
         \textcolor{green}{<|endofbelief|>} \\
         
    \specialrule{.3em}{.2em}{.2em}
    
    \end{tabular}
    \vspace{1.5mm}
    \caption{Examples of Type 1 noisy-labeled annotation, context lacks enough information to infer the true belief state.}
    \label{tab:appendix-type1-annotation}
\end{table}
\vspace{10mm}
\newpage

\begin{table}[htb!]
    \centering
    \scriptsize
    \begin{tabular}{c|p{10cm}}
    \specialrule{.3em}{.2em}{.2em}

    \multirow{3}{*}{\textbf{Context ('MUL0088', turn 2)}} & {\textcolor{blue}{<|context|>} \textcolor{TealBlue}{<|user|>}   
    i am looking for a cheap hotel with free parking near cambridge . \textcolor{Periwinkle}{<|system|>} i have multiple cheap hotel -s with free parking . what part of town are you interested in staying in ? \textcolor{TealBlue}{<|user|>} \hl{i would like to stay close to the center area , but the hotel should be 3 star} .
    \textcolor{blue}{<|endofcontext|>}}
 \\
        \hline
        \multirow{1}{*}{\textbf{Target}}  & \textcolor{green}{<|belief|>} 
        hotel parking yes, hotel pricerange cheap
        \textcolor{green}{<|endofbelief|>} 
 \\
        \hline
        \multirow{1}{*}{\textbf{SimpleTOD}} & \textcolor{green}{<|belief|>}
        \hl{hotel area centre}, hotel parking yes, hotel pricerange cheap, \hl{hotel stars 3}
         \textcolor{green}{<|endofbelief|>} 
\\
    \specialrule{.3em}{.2em}{.2em}
    
      \multirow{5}{*}{\textbf{Context ('PMUL2437', turn 3)}} & {\textcolor{blue}{<|context|>} \textcolor{TealBlue}{<|user|>} i want to find a moderate -ly priced restaurant . \textcolor{Periwinkle}{<|system|>} i have many options available for you ! is there a certain area or cuisine that interests you ? \textcolor{TealBlue}{<|user|>} yes i would like the restaurant to be located in the center of the attractions . \textcolor{Periwinkle}{<|system|>} there are 21 restaurant -s available in the centre of town . how about a specific type of cuisine ? \textcolor{TealBlue}{<|user|>} i need to know the food type and postcode and it should also have \hl{mutliple sports} <|endofcontext|>}
 \\
        \hline
        \multirow{1}{*}{\textbf{Target}}  & \textcolor{green}{<|belief|>} restaurant area centre, restaurant pricerange moderate \textcolor{green}{<|endofbelief|>} 
 \\
        \hline
        \multirow{2}{*}{\textbf{SimpleTOD}} & \textcolor{green}{<|belief|>} \hl{attraction type multiple sports}, restaurant area centre, restaurant pricerange moderate \textcolor{green}{<|endofbelief|>} 
\\
    
    \specialrule{.3em}{.2em}{.2em}
    
     \multirow{5}{*}{\textbf{Context ('PMUL2437', turn 3)}} & {\textcolor{blue}{<|context|>} \textcolor{TealBlue}{<|user|>} i want to find a moderate -ly priced restaurant . \textcolor{Periwinkle}{<|system|>} i have many options available for you ! is there a certain area or cuisine that interests you ? \textcolor{TealBlue}{<|user|>} yes i would like the restaurant to be located in the center of the attractions . \textcolor{Periwinkle}{<|system|>} there are 21 restaurant -s available in the centre of town . how about a specific type of cuisine ? \textcolor{TealBlue}{<|user|>} i need to know the food type and postcode and it should also have \hl{mutliple sports} <|endofcontext|>}
 \\
        \hline
        \multirow{1}{*}{\textbf{Target}}  & \textcolor{green}{<|belief|>} restaurant area centre, restaurant pricerange moderate \textcolor{green}{<|endofbelief|>} 
 \\
        \hline
        \multirow{2}{*}{\textbf{SimpleTOD}} & \textcolor{green}{<|belief|>} \hl{attraction type multiple sports}, restaurant area centre, restaurant pricerange moderate \textcolor{green}{<|endofbelief|>} 
\\
    \specialrule{.3em}{.2em}{.2em}

    \multirow{10}{*}{\textbf{Context ('MUL1060', turn 4)}} & {\textcolor{blue}{<|context|>} \textcolor{TealBlue}{<|user|>}   
    hello , i would like to find a hotel that include -s free parking . \textcolor{Periwinkle}{<|system|>} most of the hotel -s in town offer free parking . is there a certain area you would like to stay in , or do you have a price range in mind ? \textcolor{TealBlue}{<|user|>} yes . the centre would be nice and also \hl{free wifi} . \textcolor{Periwinkle}{<|system|>} the university arms is an expensive , 4 star hotel with free wifi . comparatively , the alexander bed and breakfast is a cheap -ly priced guesthouse , also 4 stars . \textcolor{TealBlue}{<|user|>} please book me some rooms for the university arms to accommodate 8 people for 3 nights starting on wednesday . can you also provide me the reference number after you book ? \textcolor{Periwinkle}{<|system|>} your reference number is x5ny66zv . \textcolor{TealBlue}{<|user|>} thank you . can you please help me find a place to go in town in the same area as the hotel ? \hl{preferably a college .}
    \textcolor{blue}{<|endofcontext|>}}
 \\
        \hline
        \multirow{3}{*}{\textbf{Target}}  & \textcolor{green}{<|belief|>} 
        attraction area centre, \hl{attraction name college}, hotel area centre, hotel book day wednesday, hotel book people 8, hotel book stay 3, hotel name university arms hotel, hotel parking yes
        \textcolor{green}{<|endofbelief|>} 
 \\
        \hline
        \multirow{3}{*}{\textbf{SimpleTOD}} & \textcolor{green}{<|belief|>}
        attraction area centre, \hl{attraction type college}, hotel area centre, hotel book day wednesday, hotel book people 8, hotel book stay 3, \hl{hotel internet yes}, hotel name university arms hotel, hotel parking yes
         \textcolor{green}{<|endofbelief|>} 
\\
    \specialrule{.3em}{.2em}{.2em}
    
    \multirow{10}{*}{\textbf{Context ('MUL1642', turn 5)}} & {\textcolor{blue}{<|context|>} \textcolor{TealBlue}{<|user|>}   
    hello , i am trying to find a train that goes from cambridge to london kings cross . can you help me book a ticket ? \textcolor{Periwinkle}{<|system|>} i can help with that . can you tell me what day you will be traveling ? \textcolor{TealBlue}{<|user|>} i need to leave on saturday after 18:45 . \textcolor{Periwinkle}{<|system|>} the soonest departure time would be at 19:00 on saturday , is that okay ? \textcolor{TealBlue}{<|user|>} yes , that s perfect . can you book that for 8 people ? \textcolor{Periwinkle}{<|system|>} you are all booked with reference number 144vdbrm . the cost of 151.04 gbp will be payable at the station . can i be of further assistance today ? \textcolor{TealBlue}{<|user|>} i am looking for an expensive place to eat in the centre , what is there that fits that criteria ? \textcolor{Periwinkle}{<|system|>} there 33 place -s that fit your criteria . do you have a particular cuisine type in mind so that i can narrow the results down ? \textcolor{TealBlue}{<|user|>} \hl{it does not matter what kind of food . what would you recommend for a large group of 8 people ?}
    \textcolor{blue}{<|endofcontext|>}}
 \\
        \hline
        \multirow{2}{*}{\textbf{Target}}  & \textcolor{green}{<|belief|>} 
        restaurant area centre, restaurant pricerange expensive, train book people 8, train day saturday, train departure cambridge, train destination london kings cross, train leaveat 18:45
        \textcolor{green}{<|endofbelief|>} 
 \\
        \hline
        \multirow{3}{*}{\textbf{SimpleTOD}} & \textcolor{green}{<|belief|>}
        restaurant area centre, \hl{restaurant book people 8}, restaurant pricerange expensive, train book people 8, train day saturday, train departure cambridge, train destination london kings cross, train leaveat 18:45
         \textcolor{green}{<|endofbelief|>} 
\\
    \specialrule{.3em}{.2em}{.2em}

      \multirow{12}{*}{\textbf{Context ('MUL0088', turn 7)}} & {\textcolor{blue}{<|context|>} \textcolor{TealBlue}{<|user|>}   
    i am looking for a cheap hotel with free parking near cambridge . \textcolor{Periwinkle}{<|system|>} i have multiple cheap hotel -s with free parking . what part of town are you interested in staying in ? \textcolor{TealBlue}{<|user|>} i would like to stay close to the center area , but the hotel should be 3 star . \textcolor{Periwinkle}{<|system|>} we do not have any hotel -s that match your search . do you want to try something else ? \textcolor{TealBlue}{<|user|>} are there any moderate 3 star hotel -s with free parking ? \textcolor{Periwinkle}{<|system|>} we do not have any hotel -s that match your search . do you want to try something else ? \textcolor{TealBlue}{<|user|>} how about 1 in the moderate price range ? \textcolor{Periwinkle}{<|system|>} we have 3 entries that match your preferences . would you prefer north , south , or west ? \textcolor{TealBlue}{<|user|>} i do not have a preference but i would also like ot find a restaurant called the cow pizza kitchen and bar . \textcolor{Periwinkle}{<|system|>} before i book your restaurant would you like to book your lodging ? i think you will like \hl{hamilton lodge} . it meets your needs . \textcolor{TealBlue}{<|user|>} does it have internet ? \textcolor{Periwinkle}{<|system|>} yes , the hamilton lodge has internet . \textcolor{TealBlue}{<|user|>} sounds great . what is the address and contact information ?
    \textcolor{blue}{<|endofcontext|>}}
 \\
        \hline
        \multirow{2}{*}{\textbf{Target}}  & \textcolor{green}{<|belief|>} 
    hotel internet yes, hotel parking yes, hotel pricerange moderate, hotel stars 3, restaurant name cow pizza kitchen and bar
        \textcolor{green}{<|endofbelief|>} 
 \\
        \hline
        \multirow{2}{*}{\textbf{SimpleTOD}} & \textcolor{green}{<|belief|>}
    hotel internet yes, \hl{hotel name hamilton lodge}, hotel parking yes, hotel pricerange moderate, hotel stars 3, restaurant name cow pizza kitchen and bar
         \textcolor{green}{<|endofbelief|>} \\
         
     \specialrule{.3em}{.2em}{.2em}

  \end{tabular}
  \vspace{1.5mm}
    \caption{Examples of Type 2 noisy-labeled annotation, belief state is not labeled, while context contains the information.}
    \label{tab:appendix-type2-annotation-1}
  \end{table}

\begin{table}[htb!]
    \centering
    \scriptsize
    \begin{tabular}{c|p{10cm}}
    \specialrule{.3em}{.2em}{.2em}
    
         \multirow{2}{*}{\textbf{Context ('MUL1926', turn 1)}} & {\textcolor{blue}{<|context|>} \textcolor{TealBlue}{<|user|>}   
    yes hello ! i am looking for \hl{a train leaving on tuesday} from norwich to cambridge .
    \textcolor{blue}{<|endofcontext|>}}
 \\
        \hline
        \multirow{1}{*}{\textbf{Target}}  & \textcolor{green}{<|belief|>} 
    train departure norwich, train destination cambridge
        \textcolor{green}{<|endofbelief|>} 
 \\
        \hline
        \multirow{1}{*}{\textbf{SimpleTOD}} & \textcolor{green}{<|belief|>}
    \hl{train day tuesday}, train departure norwich, train destination cambridge
         \textcolor{green}{<|endofbelief|>} \\
     \specialrule{.3em}{.2em}{.2em}

  \multirow{5}{*}{\textbf{Context ('MUL1650', turn 3)}} & {\textcolor{blue}{<|context|>} \textcolor{TealBlue}{<|user|>}   
    i am looking for a chinese restaurant in the centre of cambridge . \textcolor{Periwinkle}{<|system|>} there are quite a few chinese restaurant -s in the area ! do you have a price range you would like to stay within ? \textcolor{TealBlue}{<|user|>} no , anything is fine . can you suggest a great 1 ? \textcolor{Periwinkle}{<|system|>} jinling noodle bar looks like a good place . \textcolor{TealBlue}{<|user|>} can i have their phone number and address ? \hl{also , are there any trains that go from peterborough to cambridge ?}
    \textcolor{blue}{<|endofcontext|>}}
 \\
        \hline
        \multirow{2}{*}{\textbf{Target}}  & \textcolor{green}{<|belief|>} 
    restaurant area centre, restaurant food chinese, restaurant name jinling noodle bar
        \textcolor{green}{<|endofbelief|>} 
 \\
        \hline
        \multirow{2}{*}{\textbf{SimpleTOD}} & \textcolor{green}{<|belief|>}
    restaurant area centre, restaurant food chinese, restaurant name jinling noodle bar, \hl{train departure peterborough, train destination cambridge}
         \textcolor{green}{<|endofbelief|>} \\
     \specialrule{.3em}{.2em}{.2em}

     \multirow{5}{*}{\textbf{Context ('MUL2376', turn 3)}} & {\textcolor{blue}{<|context|>} \textcolor{TealBlue}{<|user|>}   
    where would i find the museum of archaeology and anthropology ? \textcolor{Periwinkle}{<|system|>} the museum of archaelogy and anthropology is located at university of cambridge , downing street . their phone number is 01223333516 . do you need anything else ? \textcolor{TealBlue}{<|user|>} and the attraction type is a museum , right ? \textcolor{Periwinkle}{<|system|>} yes , \hl{it is a museum} . \textcolor{TealBlue}{<|user|>} sounds great . thank you for the info !
    \textcolor{blue}{<|endofcontext|>}}
 \\
        \hline
        \multirow{1}{*}{\textbf{Target}}  & \textcolor{green}{<|belief|>} 
    attraction name museum of archaelogy and anthropology
        \textcolor{green}{<|endofbelief|>} 
 \\
        \hline
        \multirow{2}{*}{\textbf{SimpleTOD}} & \textcolor{green}{<|belief|>}
    attraction name museum of archaelogy and anthropology, \hl{attraction type museum}
         \textcolor{green}{<|endofbelief|>} \\
     \specialrule{.3em}{.2em}{.2em}

     \multirow{5}{*}{\textbf{Context ('MUL1076', turn 9)}} & {\textcolor{blue}{<|context|>} \textcolor{TealBlue}{<|user|>}   
    yes i am looking for some hotel -s that include free parking and cheap rates in the cambridge area . \textcolor{Periwinkle}{<|system|>} there are 10 hotel -s , do you have any other preferences to narrow it down ? \textcolor{TealBlue}{<|user|>} sure . i would like a guesthouse . \textcolor{Periwinkle}{<|system|>} there are 9 options available . do you have a preference for how many stars or what area specifically ? \textcolor{TealBlue}{<|user|>} i do not care . please book 1 that you recommend . \textcolor{Periwinkle}{<|system|>} ok , \hl{how about the allenbell ?} to book a stay i'll need to know the first day you want to stay , how many days , and how many guests . \textcolor{TealBlue}{<|user|>} i need it for 8 people , and for 4 nights starting saturday . \textcolor{Periwinkle}{<|system|>} great , i have booked your stay for 8 people for 4 nights . your reference number is pb41wuot . may i help with anything else ? \textcolor{TealBlue}{<|user|>} i also would like to find a museum in the west . \textcolor{Periwinkle}{<|system|>} there are several museums there , any preferences ? \textcolor{TealBlue}{<|user|>} i would like to find the museum of science , please . \textcolor{Periwinkle}{<|system|>} i am sorry , but i am having trouble locating your requested attraction . can you please restate your request ? \textcolor{TealBlue}{<|user|>} is there a museum of science ? \textcolor{Periwinkle}{<|system|>} i am not finding 1 , sorry . \textcolor{TealBlue}{<|user|>} are there any other museums in that area ? \textcolor{Periwinkle}{<|system|>} yes there are 7 museums in the west area of town for your pleasure . is there anything else you would like to narrow them down with ? \textcolor{TealBlue}{<|user|>} the museum i want should be in the west side of town and can i have the postcode for the museum . 
    \textcolor{blue}{<|endofcontext|>}}
 \\
        \hline
        \multirow{1}{*}{\textbf{Target}}  & \textcolor{green}{<|belief|>} 
    attraction area west, attraction type museum, hotel book day saturday, hotel book people 8, hotel book stay 4, hotel parking yes, hotel pricerange cheap, hotel type guesthouse
        \textcolor{green}{<|endofbelief|>} 
 \\
        \hline
        \multirow{2}{*}{\textbf{SimpleTOD}} & \textcolor{green}{<|belief|>}
    attraction area west, attraction type museum, hotel book day saturday, hotel book people 8, hotel book stay 4, \hl{hotel name allenbell}, hotel parking yes, hotel pricerange cheap, 'hotel type guesthouse
         \textcolor{green}{<|endofbelief|>} \\
     \specialrule{.3em}{.2em}{.2em}
     
    \end{tabular}
    \vspace{1.5mm}
    \caption{Examples of Type 2 noisy-labeled annotation, belief state is not labeled, while context contains the information.}
    \label{tab:appendix-type2-annotation-2}
\end{table}

\begin{table}[htb!]
    \centering
    \scriptsize
    \begin{tabular}{c|p{10cm}}
    \specialrule{.3em}{.2em}{.2em}

        \multirow{2}{*}{\textbf{Context ('SNG1036', turn 1)}} & {\textcolor{blue}{<|context|>} \textcolor{TealBlue}{<|user|>}   
    i was not planning to stay in town tonight , but it looks like i am going to have to . can you help me find a pretty cheap room ?
    \textcolor{blue}{<|endofcontext|>}}
        \\
        \hline
        \multirow{1}{*}{\textbf{Target}}  & \textcolor{green}{<|belief|>} 
        hotel pricerange cheap, \hl{hotel type hotel}
        \textcolor{green}{<|endofbelief|>} 
        \\
        \hline
        \multirow{1}{*}{\textbf{SimpleTOD}} & \textcolor{green}{<|belief|>}
        hotel pricerange cheap
         \textcolor{green}{<|endofbelief|>} \\
     \specialrule{.3em}{.2em}{.2em}
     
     \multirow{1}{*}{\textbf{Context ('PMUL1194', turn 1)}} & {\textcolor{blue}{<|context|>} \textcolor{TealBlue}{<|user|>}   
    hello , i need information about \hl{old schools} .
    \textcolor{blue}{<|endofcontext|>}}
        \\
        \hline
        \multirow{1}{*}{\textbf{Target}}  & \textcolor{green}{<|belief|>} 
        \hl{attraction type college}
        \textcolor{green}{<|endofbelief|>} 
        \\
        \hline
        \multirow{1}{*}{\textbf{SimpleTOD}} & \textcolor{green}{<|belief|>}
        attraction name old schools
         \textcolor{green}{<|endofbelief|>} \\
     \specialrule{.3em}{.2em}{.2em}
     
     \multirow{2}{*}{\textbf{Context ('SNG0284', turn 2)}} & \textcolor{blue}{<|context|>} \textcolor{TealBlue}{<|user|>}  
     i want a train from london liverpool street arriving at 9:30 \textcolor{Periwinkle}{<|system|>} which day will you be traveling ? \textcolor{TealBlue}{<|user|>} i want to leave on thursday .
 \\
        \hline
        \multirow{2}{*}{\textbf{Target}}  & \textcolor{green}{<|belief|>}  
        train arriveby 09:30, \hl{train book people 2}, train day thursday, train departure london liverpool street
        \textcolor{green}{<|endofbelief|>} 
 \\
        \hline
        \multirow{1}{*}{\textbf{SimpleTOD}} & \textcolor{green}{<|belief|>} 
        train arriveby 09:30, train day thursday, train departure london liverpool street
\\
    
    \specialrule{.3em}{.2em}{.2em}
    
    \multirow{3}{*}{\textbf{Context ('PMUL0069', turn 2)}} & \textcolor{blue}{<|context|>} \textcolor{TealBlue}{<|user|>}  
     i am looking for some hungarian food restaurant -s near the centre , please . \textcolor{Periwinkle}{<|system|>} i am sorry there are no hungarian restaurant -s near centre . \textcolor{TealBlue}{<|user|>} what kind of expensive restaurant -s are in the center of town ?
        \\
        \hline
        \multirow{1}{*}{\textbf{Target}}  & \textcolor{green}{<|belief|>}  
    restaurant area centre, \hl{restaurant food hungarian}, restaurant pricerange expensive
        \textcolor{green}{<|endofbelief|>} 
        
        \\
        \hline
        \multirow{1}{*}{\textbf{SimpleTOD}} & \textcolor{green}{<|belief|>} 
    restaurant area centre, restaurant pricerange expensive
    \textcolor{green}{<|endofbelief|>} 
        \\
    \specialrule{.3em}{.2em}{.2em}
    
      \multirow{4}{*}{\textbf{Context ('PMUL3688', turn 3)}} & {\textcolor{blue}{<|context|>} \textcolor{TealBlue}{<|user|>}  i am looking for a place to go in the centre of town . \textcolor{Periwinkle}{<|system|>} is there any type of attraction you would like to see ? \textcolor{TealBlue}{<|user|>} any of your choice . get me the address and entrance fee \textcolor{Periwinkle}{<|system|>} the holy trinity church is located on market street . the entrance fee is free . is there anything else i can help with ? \textcolor{TealBlue}{<|user|>} i also need a train out to the airport on friday . \textcolor{blue}{<|endofcontext|>}}
 \\
        \hline
        \multirow{1}{*}{\textbf{Target}}  & \textcolor{green}{<|belief|>} attraction area centre, train day friday, \hl{train destination stansted airport}\textcolor{green}{<|endofbelief|>} 
 \\
        \hline
        \multirow{1}{*}{\textbf{SimpleTOD}} & \textcolor{green}{<|belief|>} attraction area centre, train day friday \textcolor{green}{<|endofbelief|>} 
\\
  
    \specialrule{.3em}{.2em}{.2em}

\end{tabular}
    \vspace{1.5mm}
    \caption{Examples of Type 3 noisy-labeled annotation, belief state is labeled, while context lacks the information.}
    \label{tab:appendix-type3-annotation}
\end{table}

\begin{table}[htb!]
    \centering
    \scriptsize
    \begin{tabular}{c|p{10cm}}
    \specialrule{.3em}{.2em}{.2em}

    \multirow{1}{*}{\textbf{Context ('SNG02207', turn 1)}} & {\textcolor{blue}{<|context|>} \textcolor{TealBlue}{<|user|>}   
    i need to book a tax \hl{departing from gandhi .}
    \textcolor{blue}{<|endofcontext|>}}
 \\
        \hline
        \multirow{1}{*}{\textbf{Target}}  & \textcolor{green}{<|belief|>} 
        taxi departure the gandhi
        \textcolor{green}{<|endofbelief|>} 
 \\
        \hline
        \multirow{1}{*}{\textbf{SimpleTOD}} & \textcolor{green}{<|belief|>}
        \hl{taxi departure gandhi}
         \textcolor{green}{<|endofbelief|>} 
\\
    \specialrule{.3em}{.2em}{.2em}

    
      \multirow{4}{*}{\textbf{Context ('MUL1811', turn 2)}} & {\textcolor{blue}{<|context|>} \textcolor{TealBlue}{<|user|>}  
    i am looking for something to do in town . are there any interesting attraction in the north part of town ? \textcolor{Periwinkle}{<|system|>} yes , there are several places: i see a boat , a park , \hl{and 2 swimming pools} . would 1 of these work for you ? \textcolor{TealBlue}{<|user|>} ooh , \hl{a pool would be fun . can you give me their contact info ?}     
     \textcolor{blue}{<|endofcontext|>}}
 \\
        \hline
        \multirow{1}{*}{\textbf{Target}}  & \textcolor{green}{<|belief|>} 
                attraction area north, \hl{attraction type pool}
        \textcolor{green}{<|endofbelief|>} 
 \\
        \hline
        \multirow{1}{*}{\textbf{SimpleTOD}} & \textcolor{green}{<|belief|>} 
            attraction area north, \hl{attraction type swimming pool}
        \textcolor{green}{<|endofbelief|>} 
\\
  
    \specialrule{.3em}{.2em}{.2em}
    
    \multirow{1}{*}{\textbf{Context ('PMUL0548', turn 4)}} & {\textcolor{blue}{<|context|>} \textcolor{TealBlue}{<|user|>}   
    hello , i would like some information on a restaurant . \textcolor{Periwinkle}{<|system|>} do you have a certain cuisine or area in mind ? \textcolor{TealBlue}{<|user|>} i want british food . in the east , please . \textcolor{Periwinkle}{<|system|>} \hl{the graffton hotel restaurant} is in the east and serves british food . it s location is 619 newmarket rd . fen ditton . \textcolor{TealBlue}{<|user|>} i will need it for saturday , 7 people at 17:45 please . \textcolor{Periwinkle}{<|system|>} i have successfully booked a table for 7 on saturday at 17:45 for you . the table will be reserved for 15 minutes and your reference number is r8cv0hu9 . \textcolor{TealBlue}{<|user|>} i am also looking for a train on sunday .
    \textcolor{blue}{<|endofcontext|>}}
    \\
        \hline
        \multirow{1}{*}{\textbf{Target}}  & \textcolor{green}{<|belief|>} 
    restaurant area east, restaurant book day saturday, restaurant book people 7, restaurant book time 17:45, restaurant food british, \hl{restaurant name grafton hotel restaurant}, train day sunday
        \textcolor{green}{<|endofbelief|>} 
    \\
        \hline
        \multirow{1}{*}{\textbf{SimpleTOD}} & \textcolor{green}{<|belief|>}
    restaurant area east, restaurant book day saturday, restaurant book people 7, restaurant book time 17:45, restaurant food british, \hl{restaurant name graffton hotel restaurant}, train day sunday
         \textcolor{green}{<|endofbelief|>} 
    \\
    \specialrule{.3em}{.2em}{.2em}
    
    \multirow{12}{*}{\textbf{Context ('PMUL0550', turn 7)}} & {\textcolor{blue}{<|context|>} \textcolor{TealBlue}{<|user|>}   
    hey i am looking for a train from cambridge to bishops stortford . mind helping out ? \textcolor{Periwinkle}{<|system|>} many trains are leaving on friday . mind choosing the time ? \textcolor{TealBlue}{<|user|>} i want to leave on monday and arrive by 18:45 . \textcolor{Periwinkle}{<|system|>} i have train tr4283 that leaves cambridge at 5:29 and arrives in bishops stortford at 6:07 . would you like to make reservations ? \textcolor{TealBlue}{<|user|>} could you give me the travel time and price of that train please ? \textcolor{Periwinkle}{<|system|>} the cost is 10.10 pounds . the duration of the trip is 38 minutes , leaving cambridge at 5:29 and arriving at 6:07 in bishops stortford . \textcolor{TealBlue}{<|user|>} i also want a cheap chinese restaurant . \textcolor{Periwinkle}{<|system|>} there are 4 great restaurant -s to choose from . do you prefer 1 on the south side of town or in the centre ? \textcolor{TealBlue}{<|user|>} i would like to be in the centre please . \textcolor{Periwinkle}{<|system|>} i recommend \hl{charlie chan }. would you like to reserve a table ? \textcolor{TealBlue}{<|user|>} yes . monday , 8 people , 10:30 . \textcolor{Periwinkle}{<|system|>} booking was successful . the table will be reserved for 15 minutes . reference number is : ifjmbkli . \textcolor{TealBlue}{<|user|>} thank you very much . that will be all for today . i appreciate all your help .
    \textcolor{blue}{<|endofcontext|>}}
 \\
        \hline
        \multirow{4}{*}{\textbf{Target}}  & \textcolor{green}{<|belief|>} 
restaurant area centre, restaurant book day monday, restaurant book people 8, restaurant book time 10:30, restaurant food chinese, \hl{restaurant name charlie}, restaurant pricerange cheap, train arriveby 18:45, train day monday, train departure cambridge, train destination bishops stortford
        \textcolor{green}{<|endofbelief|>} 
 \\
        \hline
        \multirow{4}{*}{\textbf{SimpleTOD}} & \textcolor{green}{<|belief|>}
restaurant area centre, restaurant book day monday, restaurant book people 8, restaurant book time 10:30, restaurant food chinese, \hl{restaurant name charlie chan}, restaurant pricerange cheap, train arriveby 18:45, train day monday, train departure cambridge, train destination bishops stortford
         \textcolor{green}{<|endofbelief|>} 
\\
    \specialrule{.3em}{.2em}{.2em}

    \end{tabular}
    \vspace{1.5mm}
    \caption{Examples of Type 4 noisy-labeled annotation, belief state value is misspelled.}
    \label{tab:appendix-type4-annotation}
\end{table}
\newpage


    
    
    
         

    

\newpage

\newpage

\begin{table}[htb!]
\vspace{-18mm}
    \centering
    \scriptsize
    \begin{tabular}{c|c|p{10cm}}
    \specialrule{.3em}{.2em}{.2em}
       \multirow{9}{*}{\textbf{Turn 1}} & Context & { \textcolor{blue}{<|context|>} \textcolor{TealBlue}{<|user|>} hello , i need a train to ely on sunday please \textcolor{blue}{<|endofcontext|>}}
 \\
        \cline{2-3}
        & \multirow{3}{*}{Target} & \textcolor{green}{<|belief|>} train destination ely \textcolor{green}{<|endofbelief|>} \textcolor{brown}{<|action|>} train inform choice, train inform day, train inform dest, train request arrive \textcolor{brown}{<|endofaction|>} \textcolor{purple}{<|response|>} we have [value\_count] trains traveling to [value\_place] on [value\_day] . what time would you like to arrive ? \textcolor{purple}{<|endofresponse|>}
 \\
        \cline{2-3}
        & \multirow{4}{*}{SimpleTOD} & \textcolor{green}{<|belief|>} train day sunday, train destination ely \textcolor{green}{<|endofbelief|>} \textcolor{brown}{<|action|>} train inform choice, train inform day, train inform leave, train inform depart, train inform dest, train inform arrive, train inform id, train offerbook none \textcolor{brown}{<|endofaction|>}  \textcolor{purple}{<|response|>} there are [value\_count] trains leaving [value\_place] on [value\_day]. [train\_id] leaves [value\_place] at [value\_time] and arrives in [value\_place] at [value\_time]. would you like to book a ticket? \textcolor{purple}{<|endofresponse|>}
    \\
    \specialrule{.3em}{.2em}{.2em}
     \multirow{10}{*}{\textbf{Turn 2}} & \multirow{3}{*}{Context} & \textcolor{blue}{<|context|>} \textcolor{TealBlue}{<|user|>} hello , i need a train to ely on sunday please \textcolor{Periwinkle}{<|system|>} we have 10 trains traveling to ely on sunday . what time would you like to arrive ? \textcolor{TealBlue}{<|user|>} how about 1 arriving by 13:00 departing from cambridge . \textcolor{blue}{<|endofcontext|>} \\
    \cline{2-3}
    & \multirow{4}{*}{Target} & \textcolor{green}{<|belief|>} train destination ely, train day sunday, train arriveby 13:00, train departure cambridge \textcolor{green}{<|endofbelief|>} \textcolor{brown}{<|action|>} train inform arrive, train inform day, train inform dest, train inform id, train offerbook none \textcolor{brown}{<|endofaction|>} \textcolor{purple}{<|response|>} [train\_id] will get you to [value\_place] by [value\_time] on [value\_day] . should i book you a ticket ? \textcolor{purple}{<|endofresponse|>} \\
    \cline{2-3}
    & \multirow{3}{*}{SimpleTOD} & \textcolor{green}{<|belief|>} train day sunday, train destination ely, train departure cambridge, train arriveby 13:00 \textcolor{green}{<|endofbelief|>} \textcolor{brown}{<|action|>} train inform arrive, train inform leave, train inform id, train offerbook none \textcolor{brown}{<|endofaction|>} \textcolor{purple}{<|response|>} [train\_id] leaves at [value\_time] and arrives at [value\_time]. would you like me to book it for you? \textcolor{purple}{<|endofresponse|>}
    \\
    \specialrule{.3em}{.2em}{.2em}
    \multirow{14}{*}{\textbf{Turn 3}} & \multirow{4}{*}{Context} & \textcolor{blue}{<|context|>} \textcolor{TealBlue}{<|user|>} hello , i need a train to ely on sunday please \textcolor{Periwinkle}{<|system|>} we have 10 trains traveling to ely on sunday . what time would you like to arrive ? \textcolor{TealBlue}{<|user|>} how about 1 arriving by 13:00 departing from cambridge . \textcolor{Periwinkle}{<|system|>} tr1159 will get you to ely by 12:07 on sunday . should i book you a ticket ? \textcolor{TealBlue}{<|user|>} yes , that would be great . i'll need tickets for 8 people . do you have a reference number ? \textcolor{blue}{<|endofcontext|>} \\
    \cline{2-3}
    & \multirow{4}{*}{Target} & \textcolor{green}{<|belief|>} train destination ely, train day sunday, train arriveby 13:00, train departure cambridge, train book people 8 \textcolor{green}{<|endofbelief|>} \textcolor{brown}{<|action|>} train offerbooked ref, train offerbooked ticket \textcolor{brown}{<|endofaction|>} \textcolor{purple}{<|response|>} booking was successful , the total fee is [value\_price] gbp payable at the station . reference number is [train\_reference] . \textcolor{purple}{<|endofresponse|>} \\
    \cline{2-3}
     & \multirow{3}{*}{SimpleTOD} & \textcolor{green}{<|belief|>} train book people 8, train arriveby 13:00, train day sunday, train departure cambridge, train destination ely \textcolor{green}{<|endofbelief|>} \textcolor{brown}{<|action|>} general reqmore none, train offerbooked ref, train offerbooked ticket \textcolor{brown}{<|endofaction|>} \textcolor{purple}{<|response|>} booking was successful, the total fee is [value\_price] gbp payable at the station. reference number is [train\_reference]. is there anything else i can help you with? \textcolor{purple}{<|endofresponse|>} \\
    \specialrule{.3em}{.2em}{.2em}
    \multirow{16}{*}{\textbf{Turn 4}} & \multirow{6}{*}{Context} & \textcolor{blue}{<|context|>} \textcolor{TealBlue}{<|user|>} hello , i need a train to ely on sunday please \textcolor{Periwinkle}{<|system|>} we have 10 trains traveling to ely on sunday . what time would you like to arrive ? \textcolor{TealBlue}{<|user|>} how about 1 arriving by 13:00 departing from cambridge . \textcolor{Periwinkle}{<|system|>} tr1159 will get you to ely by 12:07 on sunday . should i book you a ticket ? \textcolor{TealBlue}{<|user|>} yes , that would be great . i'll need tickets for 8 people . do you have a reference number ? \textcolor{Periwinkle}{<|system|>} booking was successful , the total fee is 28.16 gbp payable at the station . reference number is : hdx8ix63 . \textcolor{TealBlue}{<|user|>} i am also looking for a museum . can you recommend 1 ? \textcolor{blue}{<|endofcontext|>} \\
    \cline{2-3}
    & \multirow{3}{*}{Target} & \textcolor{green}{<|belief|>} train book people 8, train arriveby 13:00, attraction type museum, train day sunday, train departure cambridge, train destination ely \textcolor{green}{<|endofbelief|>} \textcolor{brown}{<|action|>} attraction recommend name, general greet none \textcolor{brown}{<|endofaction|>} \textcolor{purple}{<|response|>} how about [attraction\_name] ? it looks lovely \textcolor{purple}{<|endofresponse|>} \\
    \cline{2-3}
    & \multirow{4}{*}{SimpleTOD} & \textcolor{green}{<|belief|>} train book people 8, train arriveby 13:00, attraction type museum, train day sunday, train departure cambridge, train destination ely \textcolor{green}{<|endofbelief|>} \textcolor{brown}{<|action|>} attraction recommend name, attraction recommend area, attraction recommend fee, attraction recommend addr, attraction recommend phone \textcolor{brown}{<|endofaction|>} \textcolor{purple}{<|response|>} i would recommend the [attraction\_name] in the [value\_area]. it is located at [attraction\_address]. the entrance fee is [value\_count] pounds. the phone number is [attraction\_phone] \textcolor{purple}{<|endofresponse|>}
    \\
    \specialrule{.3em}{.2em}{.2em}
    \multirow{16}{*}{\textbf{Turn 5}} & \multirow{6}{*}{Context} & \textcolor{blue}{<|context|>} \textcolor{TealBlue}{<|user|>} hello , i need a train to ely on sunday please \textcolor{Periwinkle}{<|system|>} we have 10 trains traveling to ely on sunday . what time would you like to arrive ? \textcolor{TealBlue}{<|user|>} how about 1 arriving by 13:00 departing from cambridge . \textcolor{Periwinkle}{<|system|>} tr1159 will get you to ely by 12:07 on sunday . should i book you a ticket ? \textcolor{TealBlue}{<|user|>} yes , that would be great . i'll need tickets for 8 people . do you have a reference number ? \textcolor{Periwinkle}{<|system|>} booking was successful , the total fee is 28.16 gbp payable at the station . reference number is : hdx8ix63 . \textcolor{TealBlue}{<|user|>} i am also looking for a museum . can you recommend 1 ? \textcolor{Periwinkle}{<|system|>} how about cafe jello gallery ? it looks lovely \textcolor{TealBlue}{<|user|>} that will work . could you get me the phone number and area its in ? \textcolor{blue}{<|endofcontext|>}
       \\
      \cline{2-3}
     & \multirow{4}{*}{Target} & \textcolor{green}{<|belief|>} train book people 8, train arriveby 13:00, attraction type museum, train day sunday, train departure cambridge, train destination ely \textcolor{green}{<|endofbelief|>} \textcolor{brown}{<|action|>} general greet none, general reqmore none, attraction inform phone, attraction inform area \textcolor{brown}{<|endofaction|>} \textcolor{purple}{<|response|>} sure ! it s in the [value\_area] area of town , and the phone number is [attraction\_phone] . can i help you with anything else ? \textcolor{purple}{<|endofresponse|>} \\
     \cline{2-3}
    & \multirow{3}{*}{SimpleTOD} & \textcolor{green}{<|belief|>} train book people 8, train arriveby 13:00, attraction type museum, train day sunday, train departure cambridge, train destination ely \textcolor{green}{<|endofbelief|>} \textcolor{brown}{<|action|>} attraction inform phone, attraction inform area, attraction inform name\textcolor{brown}{<|endofaction|>} \textcolor{purple}{<|response|>} [attraction\_name] is in the [value\_area] area and the phone number is [attraction\_phone] \textcolor{purple}{<|endofresponse|>} \\
    \specialrule{.3em}{.2em}{.2em}
    \multirow{16}{*}{\textbf{Turn 6}} & \multirow{7}{*}{Context} & \textcolor{blue}{<|context|>} \textcolor{TealBlue}{<|user|>} hello , i need a train to ely on sunday please \textcolor{Periwinkle}{<|system|>} we have 10 trains traveling to ely on sunday . what time would you like to arrive ? \textcolor{TealBlue}{<|user|>} how about 1 arriving by 13:00 departing from cambridge . \textcolor{Periwinkle}{<|system|>} tr1159 will get you to ely by 12:07 on sunday . should i book you a ticket ? \textcolor{TealBlue}{<|user|>} yes , that would be great . i'll need tickets for 8 people . do you have a reference number ? \textcolor{Periwinkle}{<|system|>} booking was successful , the total fee is 28.16 gbp payable at the station . reference number is : hdx8ix63 . \textcolor{TealBlue}{<|user|>} i am also looking for a museum . can you recommend 1 ? \textcolor{Periwinkle}{<|system|>} how about cafe jello gallery ? it looks lovely \textcolor{TealBlue}{<|user|>} that will work . could you get me the phone number and area its in ? \textcolor{Periwinkle}{<|system|>} sure ! it s in the west area of town , and the phone number is 01223312112 . can i help you with anything else ? \textcolor{TealBlue}{<|user|>} you were great today . thanks so much for all the help . that is all . goodbye . \textcolor{blue}{<|endofcontext|>} \\
    \cline{2-3}
    & \multirow{4}{*}{Target} & \textcolor{green}{<|belief|>} train book people 8, train arriveby 13:00, attraction type museum, train day sunday, train departure cambridge, train destination ely <|endofbelief|> \textcolor{brown}{<|action|>} general bye none, general welcome none, general greet none \textcolor{brown}{<|endofaction|>} \textcolor{purple}{<|response|>} it was a pleasure to help . have a good day . goodbye . \textcolor{purple}{<|endofresponse|>} \\
    \cline{2-3}
    & \multirow{2}{*}{SimpleTOD} & \textcolor{green}{<|belief|>} train book people 8, train arriveby 13:00, attraction type museum, train day sunday, train departure cambridge, train destination ely \textcolor{green}{<|endofbelief|>} \textcolor{brown}{<|action|>} general bye none \textcolor{brown}{<|endofaction|>} \textcolor{purple}{<|response|>} thank you for using our service \textcolor{purple}{<|endofresponse|>}\\

    \specialrule{.3em}{.2em}{.2em}
    \end{tabular}
    \vspace{1mm}
    \caption{{\small SimpleTOD end-to-end generation (delexicalized) on MultiWOZ (dialogue 'PMUL3663')}. 
    }
    \label{tab:target_response_example}
\end{table}

\begin{table}[htb!]
\vspace{-18mm}
    \centering
    \scriptsize
    \begin{tabular}{c|c|p{10cm}}
    \specialrule{.3em}{.2em}{.2em}
       \multirow{10}{*}{\textbf{Turn 1}} & Context & { \textcolor{blue}{<|context|>} \textcolor{TealBlue}{<|user|>} hello , i need a train to ely on sunday please \textcolor{blue}{<|endofcontext|>}}
 \\
        \cline{2-3}
        & \multirow{4}{*}{Target}  & \textcolor{green}{<|belief|>} train destination ely \textcolor{green}{<|endofbelief|>} \textcolor{brown}{<|action|>} train inform choice, train inform day, train inform dest, train request arrive \textcolor{brown}{<|endofaction|>} \textcolor{purple}{<|response|>}  we have 10 trains traveling to ely on sunday . what time would you like to arrive ? \textcolor{purple}{<|endofresponse|>}
 \\
        \cline{2-3}
        & \multirow{4}{*}{SimpleTOD} & \textcolor{green}{<|belief|>} train day sunday, train destination ely \textcolor{green}{<|endofbelief|>} \textcolor{brown}{<|action|>} train inform choice, train inform day, train inform leave, train inform depart, train inform dest, train inform arrive, train inform id, train offerbook none \textcolor{brown}{<|endofaction|>} \textcolor{purple}{<|response|>} there are 10 trains leaving cambridge on sunday. tr3177 leaves cambridge at 9:50 and arrives in ely at 10:07. would you like to book a ticket? \textcolor{purple}{<|endofresponse|>}
    \\
    \hline
     \multirow{10}{*}{\textbf{Turn 2}} & \multirow{3}{*}{Context} & \textcolor{blue}{<|context|>} \textcolor{TealBlue}{<|user|>} hello , i need a train to ely on sunday please \textcolor{Periwinkle}{<|system|>} we have 10 trains traveling to ely on sunday . what time would you like to arrive ? \textcolor{TealBlue}{<|user|>} how about 1 arriving by 13:00 departing from cambridge . \textcolor{blue}{<|endofcontext|>} \\
    \cline{2-3}
    & \multirow{4}{*}{Target} & \textcolor{green}{<|belief|>} train destination ely, train day sunday, train arriveby 13:00, train departure cambridge \textcolor{green}{<|endofbelief|>} \textcolor{brown}{<|action|>} train inform arrive, train inform day, train inform dest, train inform id, train offerbook none \textcolor{brown}{<|endofaction|>} \textcolor{purple}{<|response|>}tr1159 will get you to ely by 12:07 on sunday . should i book you a ticket ? \textcolor{purple}{<|endofresponse|>} \\
    \cline{2-3}
    & \multirow{3}{*}{SimpleTOD} & \textcolor{green}{<|belief|>} train day sunday, train destination ely, train departure cambridge, train arriveby 13:00 \textcolor{green}{<|endofbelief|>} \textcolor{brown}{<|action|>} train inform arrive, train inform leave, train inform id, train offerbook none \textcolor{brown}{<|endofaction|>} \textcolor{purple}{<|response|>} tr1159 leaves at 11:50 and arrives at 12:07. would you like me to book it for you? \textcolor{purple}{<|endofresponse|>}
    \\
    \specialrule{.3em}{.2em}{.2em}
    \multirow{14}{*}{\textbf{Turn 3}} & \multirow{4}{*}{Context} & \textcolor{blue}{<|context|>} \textcolor{TealBlue}{<|user|>} hello , i need a train to ely on sunday please \textcolor{Periwinkle}{<|system|>} we have 10 trains traveling to ely on sunday . what time would you like to arrive ? \textcolor{TealBlue}{<|user|>} how about 1 arriving by 13:00 departing from cambridge . \textcolor{Periwinkle}{<|system|>} tr1159 will get you to ely by 12:07 on sunday . should i book you a ticket ? \textcolor{TealBlue}{<|user|>} yes , that would be great . i'll need tickets for 8 people . do you have a reference number ? \textcolor{blue}{<|endofcontext|>} \\
    \cline{2-3}
    & \multirow{4}{*}{Target} & \textcolor{green}{<|belief|>} train destination ely, train day sunday, train arriveby 13:00, train departure cambridge, train book people 8 \textcolor{green}{<|endofbelief|>} \textcolor{brown}{<|action|>} train offerbooked ref, train offerbooked ticket \textcolor{brown}{<|endofaction|>} \textcolor{purple}{<|response|>}booking was successful , the total fee is 28.16 gbp payable at the station . reference number is hdx8ix63. \textcolor{purple}{<|endofresponse|>}\\
    \cline{2-3}
     & \multirow{3}{*}{SimpleTOD} & \textcolor{green}{<|belief|>} train book people 8, train arriveby 13:00, train day sunday, train departure cambridge, train destination ely \textcolor{green}{<|endofbelief|>} \textcolor{brown}{<|action|>} general reqmore none, train offerbooked ref, train offerbooked ticket \textcolor{brown}{<|endofaction|>} \textcolor{purple}{<|response|>} booking was successful, the total fee is 28.16 gbp payable at the station. reference number is hdx8ix63. is there anything else i can help you with? \textcolor{purple}{<|endofresponse|>} \\
    \specialrule{.3em}{.2em}{.2em}
    \multirow{16}{*}{\textbf{Turn 4}} & \multirow{6}{*}{Context} & \textcolor{blue}{<|context|>} \textcolor{TealBlue}{<|user|>} hello , i need a train to ely on sunday please \textcolor{Periwinkle}{<|system|>} we have 10 trains traveling to ely on sunday . what time would you like to arrive ? \textcolor{TealBlue}{<|user|>} how about 1 arriving by 13:00 departing from cambridge . \textcolor{Periwinkle}{<|system|>} tr1159 will get you to ely by 12:07 on sunday . should i book you a ticket ? \textcolor{TealBlue}{<|user|>} yes , that would be great . i'll need tickets for 8 people . do you have a reference number ? \textcolor{Periwinkle}{<|system|>} booking was successful , the total fee is 28.16 gbp payable at the station . reference number is : hdx8ix63 . \textcolor{TealBlue}{<|user|>} i am also looking for a museum . can you recommend 1 ? \textcolor{blue}{<|endofcontext|>} \\
    \cline{2-3}
    & \multirow{4}{*}{Target}  & \textcolor{green}{<|belief|>} train book people 8, train arriveby 13:00, attraction type museum, train day sunday, train departure cambridge, train destination ely \textcolor{green}{<|endofbelief|>} \textcolor{brown}{<|action|>} attraction recommend name, general greet none \textcolor{brown}{<|endofaction|>} \textcolor{purple}{<|response|>} how about cafe jello gallery ? it looks lovely \textcolor{purple}{<|endofresponse|>} \\
    \cline{2-3}
    & \multirow{4}{*}{SimpleTOD} & \textcolor{green}{<|belief|>} train book people 8, train arriveby 13:00, attraction type museum, train day sunday, train departure cambridge, train destination ely \textcolor{green}{<|endofbelief|>} \textcolor{brown}{<|action|>} attraction recommend name, attraction recommend area, attraction recommend fee, attraction recommend addr, attraction recommend phone \textcolor{brown}{<|endofaction|>} \textcolor{purple}{<|response|>} i would recommend the cafe jello gallery in the west. it is located at 13 magdalene street. the entrance fee is 0 pounds. the phone number is 01223464646 \textcolor{purple}{<|endofresponse|>}
    \\
    \specialrule{.3em}{.2em}{.2em}
    
    \multirow{17}{*}{\textbf{Turn 5}} & \multirow{6}{*}{Context} & \textcolor{blue}{<|context|>} \textcolor{TealBlue}{<|user|>} hello , i need a train to ely on sunday please \textcolor{Periwinkle}{<|system|>} we have 10 trains traveling to ely on sunday . what time would you like to arrive ? \textcolor{TealBlue}{<|user|>} how about 1 arriving by 13:00 departing from cambridge . \textcolor{Periwinkle}{<|system|>} tr1159 will get you to ely by 12:07 on sunday . should i book you a ticket ? \textcolor{TealBlue}{<|user|>} yes , that would be great . i'll need tickets for 8 people . do you have a reference number ? \textcolor{Periwinkle}{<|system|>} booking was successful , the total fee is 28.16 gbp payable at the station . reference number is : hdx8ix63 . \textcolor{TealBlue}{<|user|>} i am also looking for a museum . can you recommend 1 ? \textcolor{Periwinkle}{<|system|>} how about cafe jello gallery ? it looks lovely \textcolor{TealBlue}{<|user|>} that will work . could you get me the phone number and area its in ? \textcolor{blue}{<|endofcontext|>}
       \\
      \cline{2-3}
     & \multirow{4}{*}{Target} & \textcolor{green}{<|belief|>} train book people 8, train arriveby 13:00, attraction type museum, train day sunday, train departure cambridge, train destination ely \textcolor{green}{<|endofbelief|>} \textcolor{brown}{<|action|>} general greet none, general reqmore none, attraction inform phone, attraction inform area \textcolor{brown}{<|endofaction|>} \textcolor{purple}{<|response|>} sure ! it s in the west area of town , and the phone number is 01223464646 . can i help you with anything else ? \textcolor{purple}{<|endofresponse|>} \\
     \cline{2-3}
    & \multirow{3}{*}{SimpleTOD} & \textcolor{green}{<|belief|>} train book people 8, train arriveby 13:00, attraction type museum, train day sunday, train departure cambridge, train destination ely \textcolor{green}{<|endofbelief|>} \textcolor{brown}{<|action|>} attraction inform phone, attraction inform area, attraction inform name\textcolor{brown}{<|endofaction|>} \textcolor{purple}{<|response|>} cafe jello gallery is in the west area and the phone number is 01223464646 \textcolor{purple}{<|endofresponse|>} \\
    \specialrule{.3em}{.2em}{.2em}
    \multirow{17}{*}{\textbf{Turn 6}} & \multirow{7}{*}{Context} & \textcolor{blue}{<|context|>} \textcolor{TealBlue}{<|user|>} hello , i need a train to ely on sunday please \textcolor{Periwinkle}{<|system|>} we have 10 trains traveling to ely on sunday . what time would you like to arrive ? \textcolor{TealBlue}{<|user|>} how about 1 arriving by 13:00 departing from cambridge . \textcolor{Periwinkle}{<|system|>} tr1159 will get you to ely by 12:07 on sunday . should i book you a ticket ? \textcolor{TealBlue}{<|user|>} yes , that would be great . i'll need tickets for 8 people . do you have a reference number ? \textcolor{Periwinkle}{<|system|>} booking was successful , the total fee is 28.16 gbp payable at the station . reference number is : hdx8ix63 . \textcolor{TealBlue}{<|user|>} i am also looking for a museum . can you recommend 1 ? \textcolor{Periwinkle}{<|system|>} how about cafe jello gallery ? it looks lovely \textcolor{TealBlue}{<|user|>} that will work . could you get me the phone number and area its in ? \textcolor{Periwinkle}{<|system|>} sure ! it s in the west area of town , and the phone number is 01223312112 . can i help you with anything else ? \textcolor{TealBlue}{<|user|>} you were great today . thanks so much for all the help . that is all . goodbye . \textcolor{blue}{<|endofcontext|>} \\
    \cline{2-3}
    & \multirow{4}{*}{Target} & \textcolor{green}{<|belief|>} train book people 8, train arriveby 13:00, attraction type museum, train day sunday, train departure cambridge, train destination ely \textcolor{green}{<|endofbelief|>} \textcolor{brown}{<|action|>} general bye none, general welcome none, general greet none \textcolor{brown}{<|endofaction|>} \textcolor{purple}{<|response|>} it was a pleasure to help . have a good day . goodbye . \textcolor{purple}{<|endofresponse|>} \\
    \cline{2-3}
    & \multirow{2}{*}{SimpleTOD} & \textcolor{green}{<|belief|>} train book people 8, train arriveby 13:00, attraction type museum, train day sunday, train departure cambridge, train destination ely <|endofbelief|> \textcolor{brown}{<|action|>} general bye none \textcolor{brown}{<|endofaction|>} \textcolor{purple}{<|response|>} thank you for using our service \textcolor{purple}{<|endofresponse|>}\\

    \specialrule{.3em}{.2em}{.2em}
    \end{tabular}
    \vspace{1mm}
    \caption{{\small SimpleTOD end-to-end generation (lexicalized) on MultiWOZ (dialogue 'PMUL3663')}. 
    }
    \label{tab:target_response_example_lex}
\end{table}
\bibliographystyle{abbrvnat}
\clearpage
\appendix

\end{document}